\documentclass[sigconf]{acmart}
\renewcommand\footnotetextcopyrightpermission[1]{}
\usepackage{tikz} 
\usetikzlibrary{arrows.meta, positioning, calc}
\usepackage{subcaption}
\usepackage[dvipsnames]{xcolor}
\usepackage{enumitem}
\usepackage{microtype}

\usepackage{pifont}
\usepackage{makecell}    
\usepackage{multirow}    
\usepackage{bbm}
\usepackage[normalem]{ulem}
\usepackage{booktabs}
\usepackage{colortbl}

\begin{document}

\title{DenseStep2M: A Scalable, Training-Free Pipeline for Dense Instructional Video Annotation}

\author{Mingji Ge}
\email{mingjige@sjtu.edu.cn}
\orcid{0009-0005-7102-9366}
\affiliation{%
  \institution{SAI, Shanghai Jiao Tong University}
  \state{Shanghai}
  \country{China}
}

\author{Qirui Chen}
\email{chen_qirui@sjtu.edu.cn}
\affiliation{%
  \institution{SAI, Shanghai Jiao Tong University}
  \city{Shanghai}
  \country{China}
}

\author{Zeqian Li}
\email{lzq0103@sjtu.edu.cn}
\affiliation{%
  \institution{SAI, Shanghai Jiao Tong University}
  \city{Shanghai}
  \country{China}
}

\author{Weidi Xie}
\email{weidi@sjtu.edu.cn}
\affiliation{%
  \institution{SAI, Shanghai Jiao Tong University}
  \city{Shanghai}
  \country{China}
}

\begin{abstract}
Long-term video understanding requires interpreting complex temporal events and reasoning over procedural activities. While instructional video corpora, like HowTo100M, offer rich resources for model training, they present significant challenges, including noisy ASR transcripts and inconsistent temporal alignments between narration and visual content. In this work, we introduce an automated, training-free pipeline to extract high-quality procedural annotations from in-the-wild instructional videos. Our approach segments videos into coherent shots, filters poorly aligned content, and leverages state-of-the-art multimodal and large language models (Qwen2.5-VL and DeepSeek-R1) to generate structured, temporally grounded procedural steps.

This pipeline yields \textbf{DenseStep2M}, a large-scale dataset comprising approximately 100K videos and 2M detailed instructional steps, designed to support comprehensive long-form video understanding. To rigorously evaluate our pipeline, we curate \textbf{DenseCaption100}, a benchmark of high-quality, human-written captions. Evaluations demonstrate strong alignment between our auto-generated steps and human annotations. Furthermore, we validate the utility of DenseStep2M across three core downstream tasks: dense video captioning, procedural step grounding, and cross-modal retrieval. Models fine-tuned on DenseStep2M achieve substantial gains in captioning quality and temporal localization, while exhibiting robust zero-shot generalization across egocentric, exocentric, and mixed-perspective domains. These results underscore the effectiveness of DenseStep2M in facilitating advanced multimodal alignment and long-term activity reasoning. 
Our \href{https://huggingface.co/datasets/mingjige/DenseStep2M}{dataset} is available.
\end{abstract}

\begin{teaserfigure}
    \centering
  \includegraphics[width=0.95\textwidth]{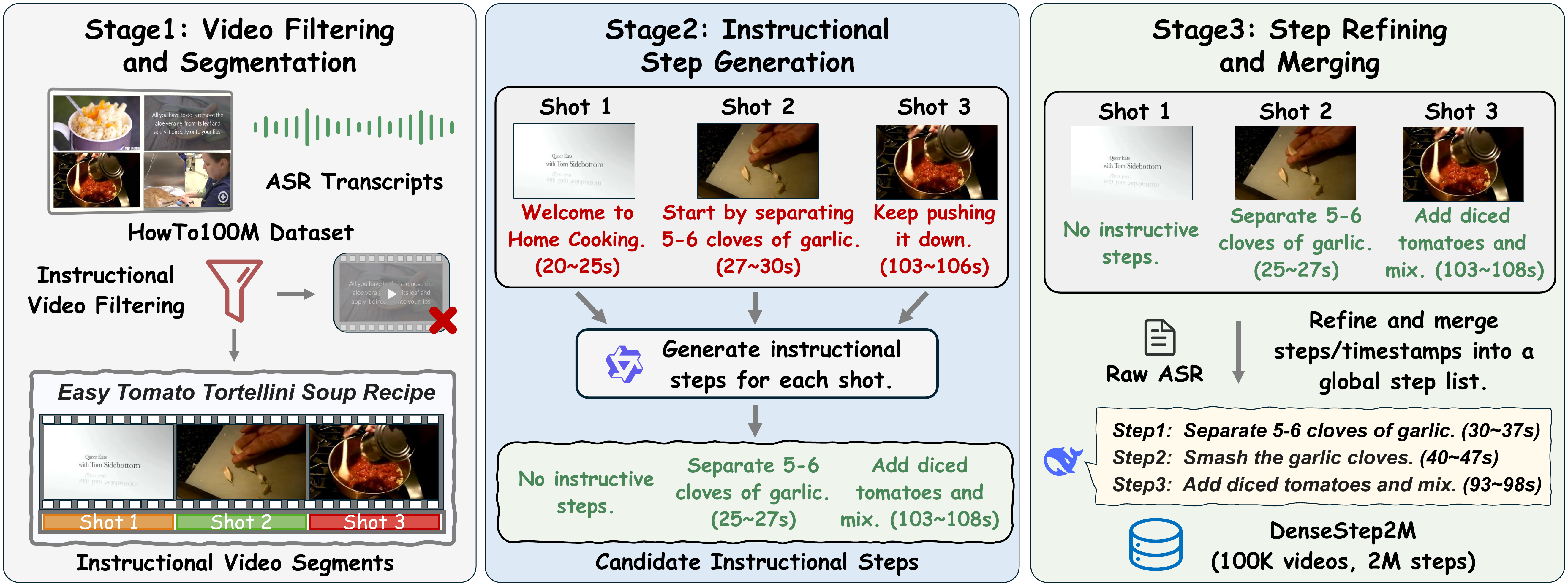}
  \caption{\textbf{The proposed three-stage pipeline for generating temporally grounded procedural steps from instructional videos.} It first filters and segments videos from HowTo100M~\cite{miech2019howto100m}, then generates candidate steps with Qwen2.5-VL-72B~\cite{bai2025qwen2}, and finally refines and merges them with DeepSeek-R1~\cite{guo2025deepseek} to construct the DenseStep2M dataset for downstream applications.}
  \Description{A three-stage pipeline diagram for generating the DenseStep2M dataset from instructional videos. Stage 1 (Video Filtering and Segmentation) shows a process where videos from the HowTo100M dataset are filtered and partitioned into sequential shots, illustrated with an "Easy Tomato Tortellini Soup Recipe" video divided into Shot 1, Shot 2, and Shot 3. Stage 2 (Instructional Step Generation) depicts the generation of candidate instructional steps for each shot. For example, Shot 2 is associated with the text "Start by separating 5-6 cloves of garlic (27-30s)," and Shot 3 is associated with "Keep pushing it down (103-106s)." Stage 3 (Step Refining and Merging) illustrates how raw ASR and candidate steps are processed to refine and merge timestamps into a global step list. The final output is a structured sequence: Step 1 (Separate garlic, 30-37s), Step 2 (Smash garlic, 40-47s), and Step 3 (Add tomatoes, 93-98s), culminating in the DenseStep2M dataset containing 100K videos and 2M steps.}
  \label{fig:Pipeline}
\end{teaserfigure}

\maketitle

\pagestyle{plain}       
\thispagestyle{empty}

\section{Introduction}

Recent advances in computer vision have spurred growing interest in \textit{long-term video understanding}~\cite{chandrasegaran2024hourvideo, mangalam2023egoschema, wu2024longvideobench, MLVU}, which involves interpreting temporally extended events~ \cite{li2024mvbench, shangguan2024tomato, ye2025re}, tracking object state changes~\cite{menon2024generating, soucek2025showhowto, xue2024learning}, 
and reasoning over procedural activities~\cite{chang2020procedure, li2020hero, song2023ego4d, wang2023pdpp, yagcioglu2018recipeqa, zhukov2019cross}.
These capabilities are central to real-world applications such as multimodal personal assistants~\cite{cheng2024videollama}, egocentric smart devices~\cite{yang2025egolifeegocentriclifeassistant}, and embodied AI systems~\cite{brohan2023rt, cheang2024gr}. 
Despite this progress, existing video-language models are predominantly trained on datasets composed of short, web-crawled clips with coarse, automatically generated descriptions~\cite{Bain21, chen2024sharegpt4video, chen2024panda, wang2023internvid, xue2022advancing, zellers2021merlot}. These datasets lack the narrative complexity, temporal coherence, and multi-shot dynamics required to train robust long-term reasoning models. In contrast, large-scale instructional video corpora—such as HowTo100M~\cite{miech2019howto100m}, offer rich procedural demonstrations and spoken guidance, making them ideal candidates for learning extended, structured activities.

However, naively leveraging in-the-wild instructional datasets poses significant challenges. First, transcripts obtained via automatic speech recognition (ASR) are notoriously noisy, often containing colloquial language, redundancies, and ambiguous references ({\em e.g.}, pronouns like `this' or `that'). Second, the temporal alignment between narrations and visual content is frequently inconsistent, with spoken instructions sometimes preceding or lagging behind the associated physical action~\cite{han2022temporal}. If not properly addressed, this semantic and temporal noise severely degrades the performance of downstream models.

Prior attempts to address these challenges have employed early vision-language models and autoregressive large language models (LLMs)~\cite{dvornik2023stepformer, li2024multi, shvetsova2024howtocaption}. Unfortunately, the vision models of that era lacked the fine-grained visual perception necessary to accurately align textual instructions with subtle visual cues. Similarly, early LLMs were constrained by limited context windows, preventing the holistic processing of lengthy ASR transcripts. These model-level limitations resulted in fragmented temporal reasoning, frequently leading to imprecise step localization and hallucinated or semantically irrelevant procedural descriptions.

Recent breakthroughs in frontier multimodal large language models (MLLMs) and reasoning-focused LLMs provide an unprecedented opportunity to overcome these bottlenecks. Here, we introduce a highly scalable, automated, and training-free pipeline to extract high-quality procedural annotations from noisy instructional videos. Our pipeline first segments long-form videos into visually coherent shots and filters out segments with poor visual-textual alignment. We then employ state-of-the-art models, specifically Qwen2.5-VL-72B~\cite{bai2025qwen2} and DeepSeek-R1-671B~\cite{guo2025deepseek}, to jointly reason over video frames and transcripts. This synergistic approach allows us to generate structured, temporally grounded procedural steps, that are precisely aligned with specific video segments.

Applying this pipeline yields \textbf{DenseStep2M}, a large-scale dataset comprising approximately 100K instructional videos annotated with 2 million fine-grained, semantically rich procedural steps. To rigorously validate our pipeline and the resulting dataset, we curate \textbf{DenseCaption100}, a high-quality benchmark of 100 instructional videos with dense, human-written annotations. Evaluations demonstrate that our automated annotations achieve strong alignment with human standards. Furthermore, fine-tuning vision-language models on DenseStep2M yields substantial performance gains across both generative~(dense captioning and procedural grounding) and discriminative~(cross-modal retrieval) video understanding tasks.

\section{Related Work}

\vspace{3pt}\noindent \textbf{Large-Scale Video-Text Datasets.}
Multimodal video-text data\-sets serve as fundamental resources for advancing video understanding research. Early datasets such as MSVD~\cite{chen2011collecting}, MSR-VTT~\cite{xu2016msr}, and DiDeMo~\cite{anne2017localizing} provide high-quality, human-curated annotations but are limited in scale, often containing only a few thousand clips. To support the training of foundational models, the field has shifted toward web-scale corpora. 
Datasets like WebVid-2M~\cite{bain2021frozen}, Youtube-8M~\cite{abu2016youtube}, and Instagram65M~\cite{ghadiyaram2019large} leverage massive amounts of internet data, though they often rely on sparse user-provided tags or short captions. More recently, Panda-70M~\cite{chen2024panda} and InternVid~\cite{wang2023internvid} have utilized cross-modality teachers to generate millions of detailed descriptions, while HD-VILA-100M~\cite{xue2022advancing}  focuses on high-resolution video-language alignment through large-scale transcriptions. However, these datasets are predominantly composed of short clips with coarse-grained annotations. 
Even large-scale instructional datasets like HowTo100M~\cite{miech2019howto100m} that provide rich procedural content, suffer from significant noise in their ASR transcripts and temporal misalignment between narration and visual action. 
These limitations hinder their effectiveness for training models that require multi-event, long-term video understanding.

\vspace{3pt}\noindent \textbf{Instructional Video Datasets.}
Instructional videos demonstrate the procedural steps for completing specific tasks, wherein the presenter explicates the actions, objects, and other elements involved in each step, providing rich multimodal information encompassing visual, auditory, and textual content~\cite{alayrac2016unsupervised,regneri2013grounding}.
Existing instructional video-text datasets~\cite{afouras2023ht, kuehne2014language, song2023ego4d, tang2019coin, zala2023hierarchical, zhou2018towards, zhukov2019cross}, including third-person and egocentric perspectives, have been widely utilized in prior research.
However, these datasets typically rely on manually defined, closed-set taxonomies of procedural steps, and many are restricted to the cooking domain.
To increase dataset scale and task diversity, recent studies~\cite{dvornik2023stepformer,li2024multi, mavroudi2023learning, shvetsova2024howtocaption, soucek2025showhowto} have proposed automated methods for extracting procedural video-text data from large-scale in-the-wild video datasets, {\em e.g.,} the HowTo100M dataset~\cite{miech2019howto100m}.
The resulting large-scale open-vocabulary instructional video-text datasets have multiple potential applications, such as procedural question answering~\cite{yagcioglu2018recipeqa}, dense video captioning~\cite{zhou2018towards}, object state change understanding~\cite{xue2024learning}, and instruction-conditioned image generation~\cite{menon2024generating, souvcek2024genhowto}.

\vspace{3pt}\noindent \textbf{Multimodal Large Language Models.}
Recent advances in multimodal large language models (MLLMs) have demonstrated remarkable capabilities in visual understanding and reasoning, with open-source models increasingly narrowing the performance gap with proprietary systems.
Notably, open-source models such as Qwen-VL series~\cite{bai2023qwen, wang2024qwen2,qwen2.5,li2026qwen3}, Intern-VL series~\cite{chen2024internvl,chen2024expanding,chen2024far,zhu2025internvl3} have already achieved performance comparable to commercial models~\cite{Anthropic2024Claude3,hurst2024gpt,singh2025openai,team2023gemini,comanici2025gemini}, laying the foundation for large-scale data annotation and refinement tasks that were previously reliant on expensive API-based solutions.
Building upon these developments, we construct an automated pipeline utilizing multiple advanced open-source MLLMs and reasoning LLMs to analyze and denoise video-text datasets, achieving significant quality improvements over prior datasets while maintaining efficiency and scalability.

\begin{figure*}[t]
    \centering
    \includegraphics[width=\textwidth]{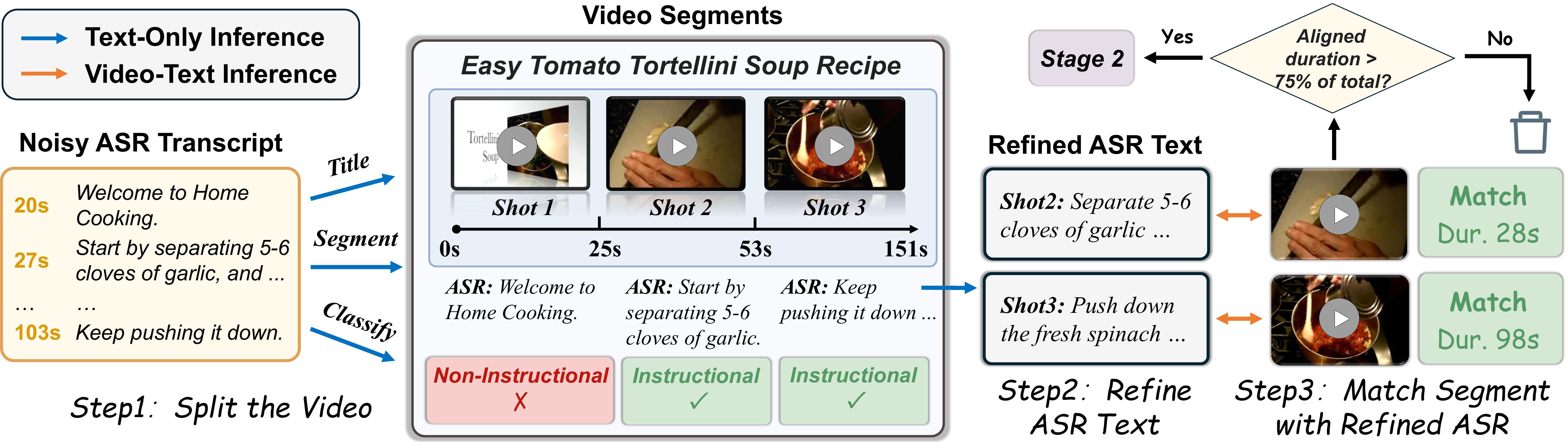} 
    \caption{\textbf{Three-step pipeline for Stage1 (Video Filtering
and Segmentation).} WhisperX~\cite{bain2022whisperx} first generates noisy ASR transcripts with timestamps, followed by Qwen2.5~\cite{qwen2.5} for titling, segmentation, and coarse instructional filtering. DeepSeek-R1~\cite{guo2025deepseek} then refines the ASR text, and Qwen2.5-VL~\cite{bai2025qwen2} finally verifies semantic alignment between video segments and the refined text. Videos with $\geq$75\% aligned instructional content are retained.}
    \Description{A detailed flowchart of Stage 1, illustrating the three-step pipeline for Video Filtering and Segmentation. Step 1 (Split the Video) shows a "Noisy ASR Transcript" being processed via "Text-Only Inference" to generate a "Title," segment the video into "Shot 1, Shot 2, and Shot 3," and "Classify" each as "Instructional" or "Non-Instructional." For example, Shot 1 is marked as Non-Instructional (X), while Shots 2 and 3 are marked as Instructional (checkmark). Step 2 (Refine ASR Text) depicts the conversion of noisy ASR into "Refined ASR Text" for specific segments, such as refining Shot 2 to "Separate 5-6 cloves of garlic." Step 3 (Match Segment with Refined ASR) utilizes "Video-Text Inference" to verify the semantic alignment between the visual segments and the refined text. It displays duration matches, such as "Match Dur. 28s" for Shot 2 and "Match Dur. 98s" for Shot 3. The pipeline concludes with a decision diamond: "Aligned duration > 75\% of total?" If the condition is met (Yes), the video proceeds to Stage 2; if not (No), the video is discarded, indicated by a trash bin icon.}
    \label{fig:SelectVideos}
\end{figure*}

\section{Automated Annotation Pipeline} 
\label{sec:dataset}

In this section, we propose an automated pipeline for generating temporally grounded procedural steps from instructional videos, using visual content and ASR transcripts as the primary signals (Sec.~\ref{subsec:Problem Formulation}). As shown in Figure~\ref{fig:Pipeline}, the pipeline is structured into three stages: 1) \textbf{Video filtering and segmentation} (Sec.~\ref{subsec:Video Filtering and Segmentation}), where (V)LLMs leverage both visual and textual inputs to retain instructional videos from the HowTo100M dataset~\cite{miech2019howto100m} and segment them into sequential shots; 2) \textbf{Instructional step generation} (Sec.~\ref{subsec:Instructional Step Generation}), where Qwen2.5-VL-72B~\cite{bai2025qwen2} produces candidate steps for each video shot by integrating visual and textual cues; and 3) \textbf{Step refinement and merging} (Sec.~\ref{subsec:Step Refining and Merging}), where DeepSeek-R1~\cite{guo2025deepseek} refines these candidate steps by improving both semantic clarity and temporal alignment, and subsequently merges them to form the complete set of procedural steps for each video. Applying this pipeline results in a large-scale \textbf{DenseStep2M} dataset (Sec.~\ref{subsec:Dataset Analysis}), which can be used to fine-tune VLLMs on various downstream tasks.

\subsection{Problem Formulation}
\label{subsec:Problem Formulation}

Given an untrimmed instructional video~($\mathcal{V}$) and its transcripts~($\mathcal{A}$) 
obtained via automatic speech recognition~(ASR), where 
$\mathcal{A} = \{(a_1, s_1, e_1), \dots, (a_m, s_m, e_m)\}$ denotes a sequence of $m$ 
sentences uttered by the demonstrator with their respective start and end timestamps 
$(s_i, e_i)$, our goal is to leverage various off-the-shelf tools to produce a structured set of instruction steps $\mathcal{C} = \{(c_1, s_1, e_1), \dots, \allowbreak (c_n, s_n, e_n)\}$ that comprehensively captures all procedural steps demonstrated in the video, with each step~$c_i$ precisely aligned to its corresponding temporal segment $(s_i, e_i)$.

\subsection{Stage1: Video Filtering and Segmentation} \label{subsec:Video Filtering and Segmentation}

The HowTo100M dataset~\cite{miech2019howto100m} offers a diverse collection of videos, yet quality varies significantly across its content: (i) many videos are non-instructional, featuring content such as travel vlogs or product showcases; (ii) even within instructional videos, non-instructional shots, such as personal anecdotes or greetings are prevalent~(see the Appendix for further details); (iii) the raw ASR transcripts, generated via the YouTube API~\cite{miech2019howto100m}, suffer from noise due to ASR system imperfections~\cite{han2022temporal}.

Our goal is to filter the HowTo100M dataset to retain high-quality instructional videos, defined as those in which at least 75\% of the total duration consists of instructional content. This is accomplished through a three-step procedure 
(Figure~\ref{fig:SelectVideos}), described below.

\vspace{3pt}\noindent \textbf{Step1: Text-based Filtering and Segmentation.}
% \paragraph{\textbf{Step1: Text-based Filtering and Segmentation.}} 
To improve ASR transcript quality, we first replace the original YouTube transcripts with those generated by WhisperX~\cite{bain2022whisperx}, a strong speech recognition model that produces accurate text with precise word-level timestamps. We then prompt 
Qwen2.5-72B~\cite{qwen2.5} to \textbf{generate a title} for each video, \textbf{segment} 
it into sequential shots, and \textbf{classify} whether each segment contains instructional content, all based solely on the ASR transcripts. Videos are retained only if instructional segments account for at least 75\% of the total duration.

However, relying solely on textual cues can be insufficient, as visual content may not 
always align with the spoken transcripts. To address this, we incorporate pre-trained 
multimodal LLMs (MLLMs) in the following two steps.

\vspace{3pt}\noindent \textbf{Step2: ASR Text Refinement.}
The noisy ASR transcripts are further refined using the reasoning model DeepSeek-R1~\cite{guo2025deepseek} to \textbf{extract key information}. Since refinement inevitably discards certain details present in the original transcripts, the subsequent Stage 2 and Stage 3 continue to operate on \textbf{the raw ASR transcripts} rather than the refined ones.

\vspace{3pt}\noindent \textbf{Step3: Visual-Textual Alignment Verification.}
% \paragraph{\textbf{Step3: Visual-Textual Alignment Verification.}}
We verify the semantic alignment between the refined ASR text and its corresponding video segments using Qwen2.5-VL-7B~\cite{bai2025qwen2}. Segments demonstrating sufficient semantic correspondence are identified as containing authentic instructional content. To ensure high data quality, only videos in which the cumulative duration of these validated segments exceeds 75\% of the total length are retained. The specific prompt templates are provided in the Appendix.

\begin{figure*}[t]
    \centering
    \includegraphics[width=0.92\textwidth]{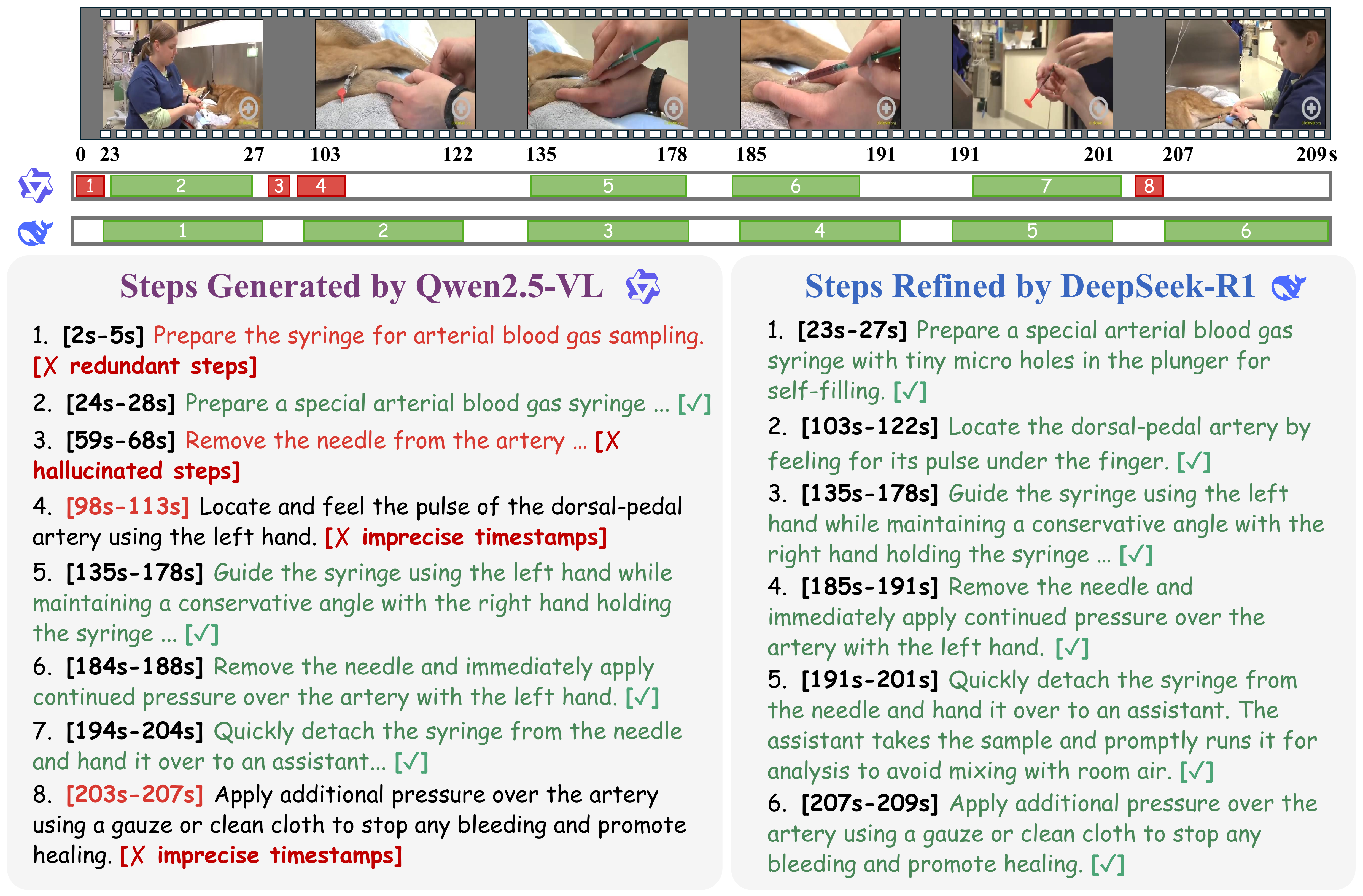}
    \caption{\textbf{Examples of issues in initial step generation by  Qwen2.5-VL}~\cite{bai2025qwen2}—such as disordered sequences, redundant or hallucinated steps, and inaccurate timestamps—addressed during Stage3 (Step Refining and Merging) by DeepSeek-R1~\cite{guo2025deepseek}.} 
    \Description{A side-by-side comparative diagram demonstrating how Stage 3 refines the initial procedural steps for a medical video depicting arterial blood gas sampling. At the top, a film strip illustrates the visual timeline of the procedure from 0 to 209 seconds. The left panel, titled "Steps Generated by Qwen2.5-VL," presents a list of 8 initial candidate steps with several marked errors. Step 1 (2-5s) is marked as a redundant step. Step 3 (59-68s) is flagged as a hallucinated action regarding needle removal. Steps 4 (98-113s) and 8 (203-207s) are labeled with imprecise timestamps. Other steps in this list feature green checkmarks but are fragmented or poorly aligned with the overall narrative. The right panel, titled "Steps Refined by DeepSeek-R1," shows a consolidated and corrected list of 6 procedural steps, all marked with green checkmarks indicating successful refinement. This refined list demonstrates superior semantic clarity and precise temporal grounding. The redundant and hallucinated steps from the initial generation are removed, and the timestamps for critical actions are rectified—for instance, the pulse location step is adjusted to 103-122s and the final pressure application to 207-209s. Two horizontal progress bars located between the film strip and the text lists visually compare the temporal spans of both annotation sets, highlighting the more coherent and continuous sequence provided by the refined version.}
    \label{fig:Steps_Refinement}
\end{figure*}

\subsection{Stage2: Instructional Step Generation} 
\label{subsec:Instructional Step Generation}

Following the segmentation in Stage 1, the video is partitioned into $N$ visually coherent shots. We define the $i$-th shot as $\mathcal{H}_i$, which encompasses the local visual content $v_i$ and its temporally aligned ASR transcript $\mathcal{A}_i$. To generate procedural annotations, we employ the Qwen2.5-VL-72B model. Our approach feeds the model with three synergistic inputs: the shot’s frames~($v_i$), the local transcript~($\mathcal{A}_i$), and the global video title~($\hat{t}$) to maintain narrative consistency.
\[ \hat{G} = \{\hat{G}_1, \dots, \hat{G}_N\}, \quad \hat{G_i} = \Phi_{\text{Qwen-2.5-VL-72B}}({v}_i; \hat{t}; \mathcal{A}_{i}), \quad \forall i \in [1, N] \]
Here, \( \hat{G_i} = \{\{g_1, s_1, e_1\}, \{g_2, s_2, e_2\}, \dots, \{g_n, s_n, e_n\}\} \) denotes the list of \( n \) instructional steps for the \( i \)-th shot, where each step \( g_j \) is paired with start and end timestamps \( s_j \) and \( e_j \). Unlike text-only ASR refinement, our use of \textbf{multimodal inputs} captures visual nuances, resulting in \textbf{more comprehensive step-level annotations} (see Sec.~\ref{subsec:Dataset Analysis}). We provide the prompt templates in the Appendix.

\subsection{Stage3: Step Refining and Merging} \label{subsec:Step Refining and Merging}
The initial steps generated in Sec.~\ref{subsec:Instructional Step Generation} often exhibit artifacts such as \textbf{disordered sequences}, \textbf{redundant or hallucinated actions}, and \textbf{imprecise temporal boundaries} (see Figure~\ref{fig:Steps_Refinement}). These deficiencies typically stem from two challenges: 
(i) \textbf{hallucinations} arising in segments with ambiguous ASR transcripts, 
and (ii) \textbf{modality conflicts}, where the model struggles to reconcile disparate visual and textual cues. While vision-language models tend to prioritize visually derived timestamps, their limited temporal grounding capacity often results in lower accuracy compared to text-based alternatives.

To mitigate these issues, we employ the reasoning-focused LLM, DeepSeek-R1~\cite{guo2025deepseek}, to perform global verification and temporal rectification. By analyzing the complete ASR transcript~($\mathcal{A}$) alongside the video title~($\hat{t}$), DeepSeek-R1 assesses the contextual plausibility of each candidate step and infers the most appropriate start and end times. This process effectively consolidates fragmented outputs from individual shots into a coherent, video-level sequence.

Formally, the refinement and merging process is defined as:
$$\hat{\mathcal{C}} = \Phi_{\text{DeepSeek-R1}}(\hat{G}; \hat{t}; \mathcal{A})$$
where $\hat{G} = \{\hat{G}_1, \dots, \hat{G}_N\}$ denotes the collection of initial shot-level step lists, and $\hat{\mathcal{C}} = \{(c_1, s_1, e_1), \dots, (c_o, s_o, e_o)\}$ represents the final set of refined steps with corrected timestamps.
Notably, although this refinement stage is purely text-based, it successfully preserves visually-identified instructional steps if they maintain \textbf{semantic consistency} within the broader ASR context. The effectiveness of this stage is demonstrated in Figure~\ref{fig:Steps_Refinement} and Table~\ref{tab:Compare architectures on DenseCaption100}, with details provided in the Appendix.

\begin{figure*}[th]
    \centering
    
    % 上子图（原第一个复杂图形）
    \begin{subfigure}[b]{\textwidth}
        \centering
        \begin{tikzpicture}[baseline=(leftimage.south)]
            % 左边图 A
            \node[anchor=south west, inner sep=0] (leftimage) at (-5.7,-0.2) {
                \includegraphics[width=0.65\textwidth]{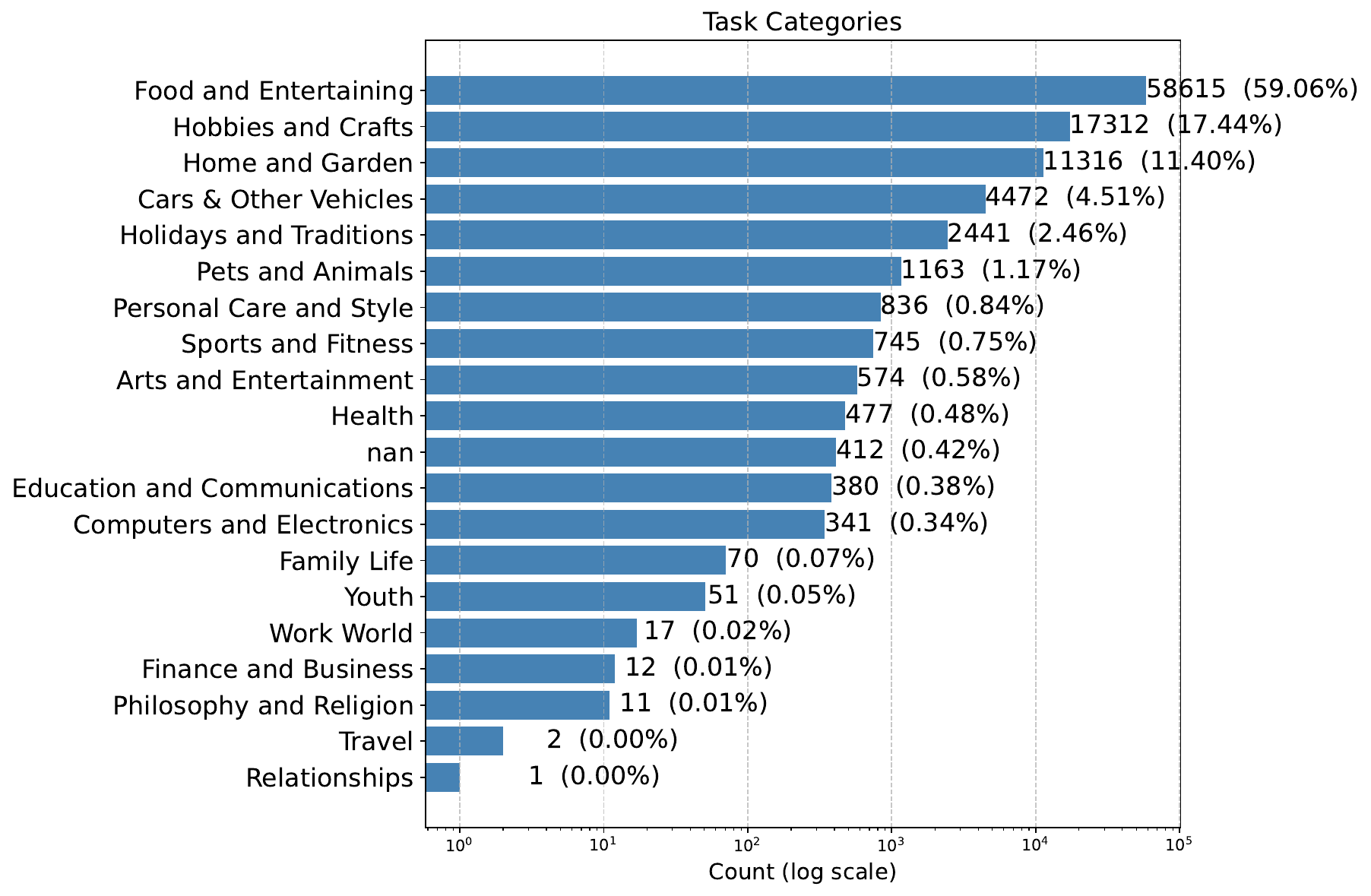}
            };
            
            % 右边图 C
            \node[anchor=south west, inner sep=0] (rightbottom) at (6,-0.2) {
                \includegraphics[width=0.35\textwidth]{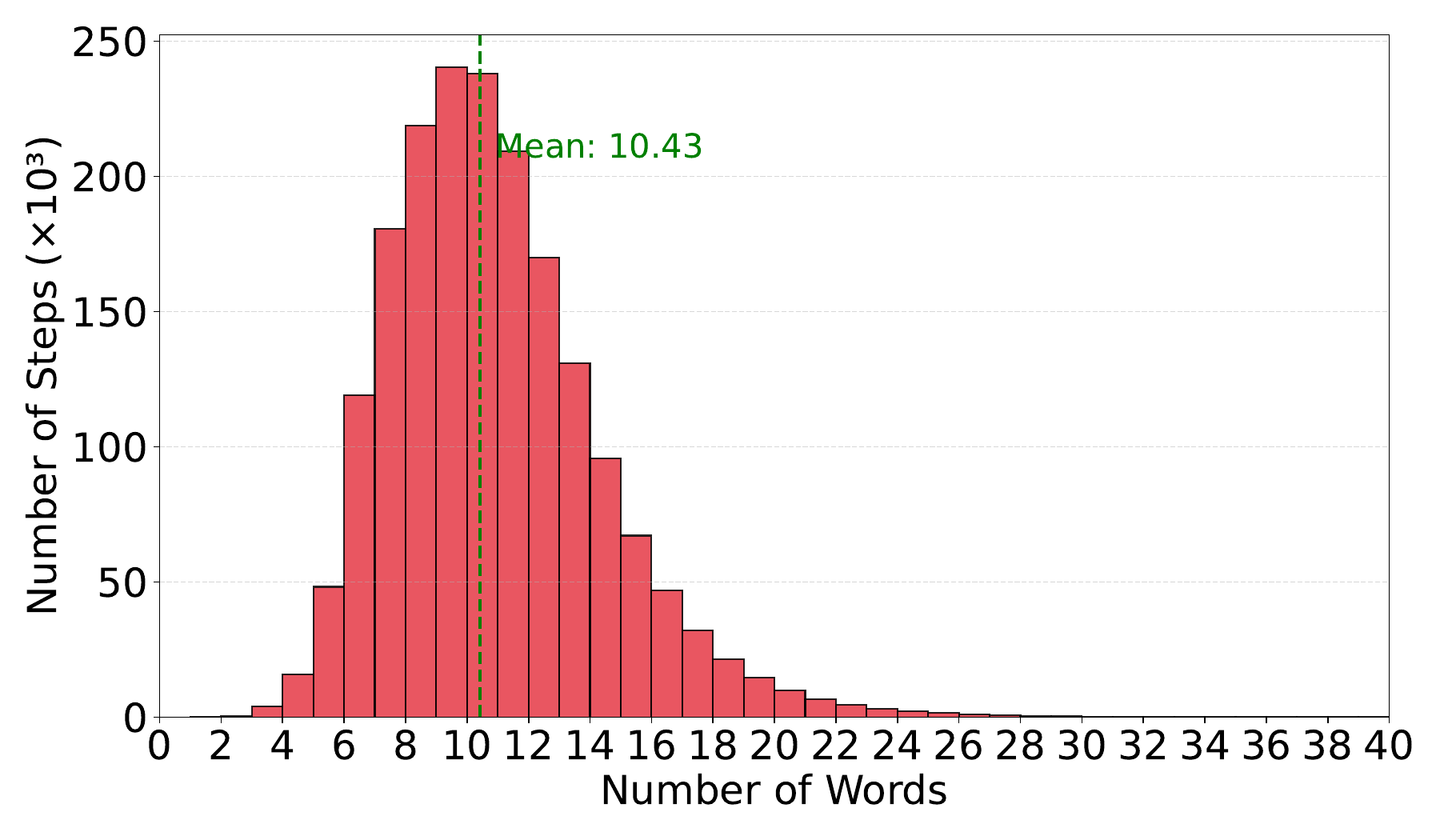}
            };
            
            % 右边图 B
            \node[anchor=south west, inner sep=0] at (6, 3.5) {
                \includegraphics[width=0.35\textwidth]{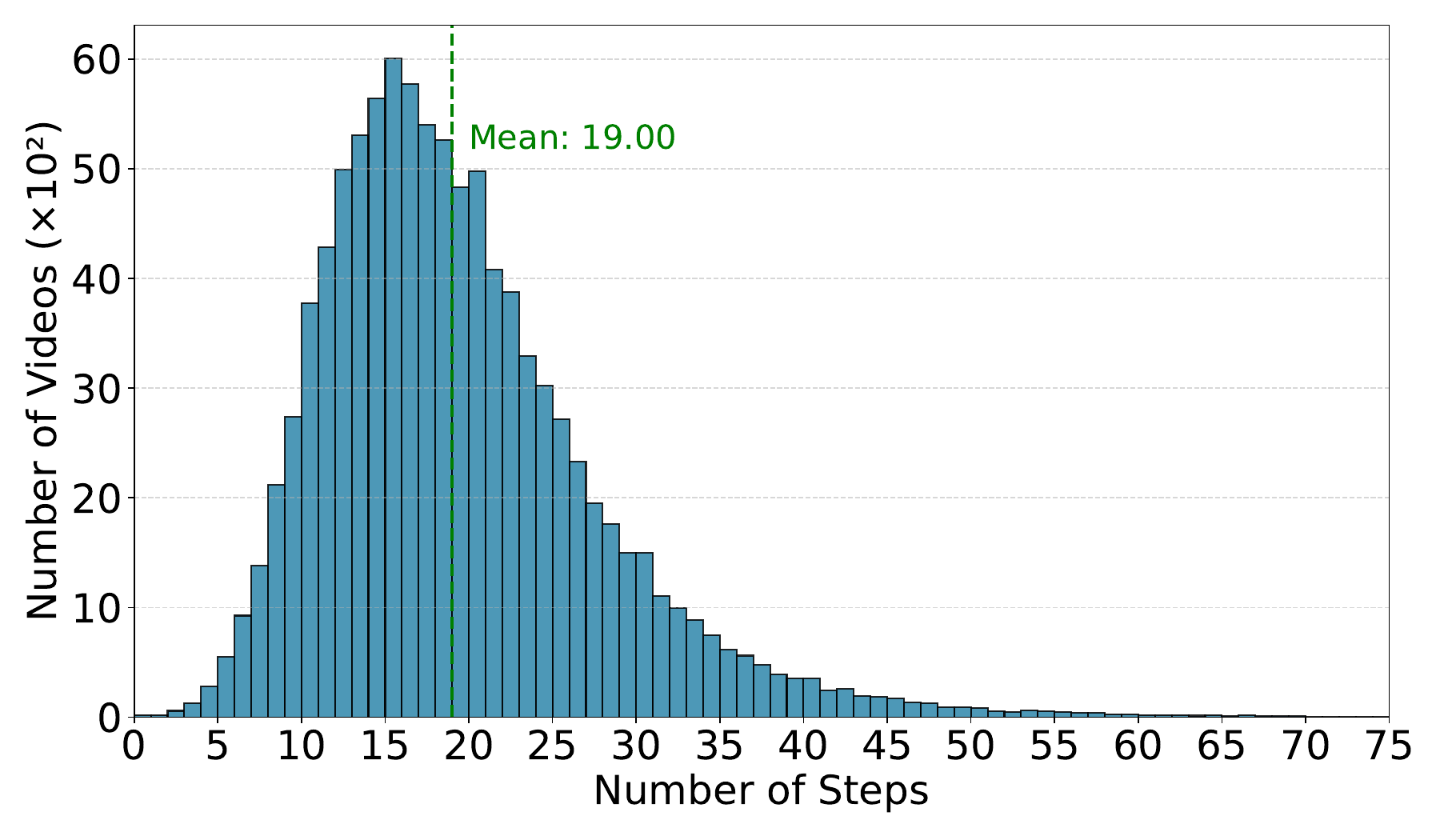}
            };
        \end{tikzpicture}
        \caption{\textbf{Statistics of our DenseStep2M dataset.}}
        \label{fig:Dataset_Analysis_sub}
    \end{subfigure}
    
    \vspace{0.3cm} % 添加垂直间距
    
    % 下子图（原第二个图形）
    \begin{subfigure}[b]{\textwidth}
        \centering
        \includegraphics[width=\textwidth]{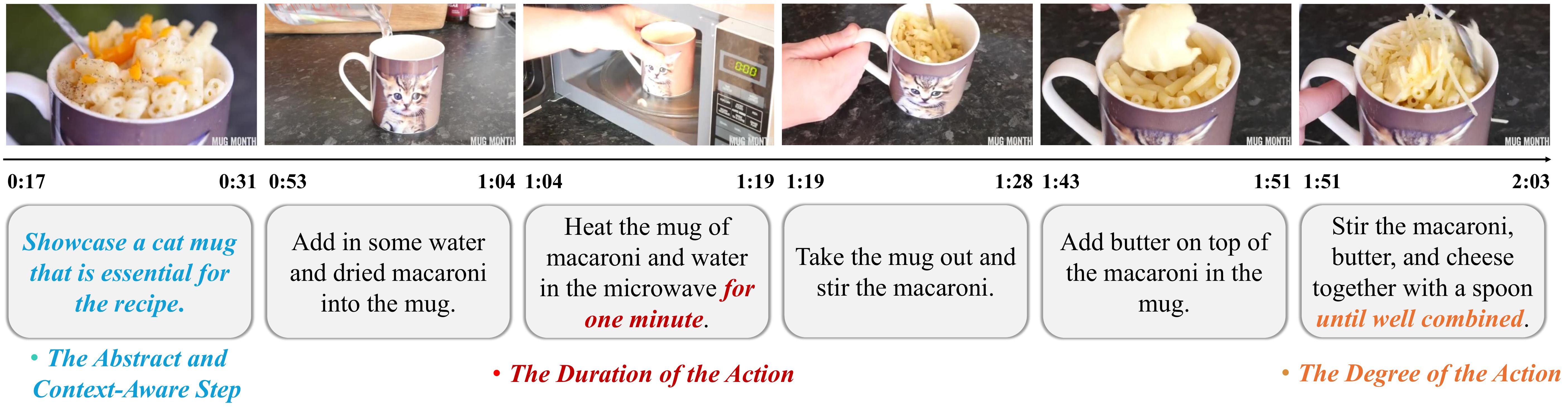}
        \caption{\textbf{Visualizations of the annotated instructional steps.}}
        \label{fig:Compare_Dataset_sub}
    \end{subfigure}
    
    \caption{\textbf{Illustrations of our DenseStep2M dataset characteristics.} (a) Distribution statistics showing task categories (left), video steps (top-right) and step words (bottom-right). (b) Annotated instructional steps showcase detailed descriptions alongside abstract and context-aware steps.}
    \Description{A two-part illustration of the DenseStep2M dataset characteristics. Part (a), titled "Statistics of our DenseStep2M dataset," contains three charts. The first is a horizontal bar chart of "Task Categories" showing that "Food and Entertaining" is the dominant category at 59.06\% (58,515 videos), followed by "Hobbies and Crafts" at 17.44\% and "Home and Garden" at 11.40\%. The top right histogram shows the distribution of the "Number of Steps" per video, with a mean of 19.00. The bottom right histogram shows the "Number of Words" per step, with a mean of 10.43. Part (b), titled "Visualizations of the annotated instructional steps," displays a timeline of six video frames and their corresponding text annotations for a macaroni recipe. The annotations highlight three unique features of the dataset: 1) "The Abstract and Context-Aware Step," exemplified by the first step "Showcase a cat mug that is essential for the recipe"; 2) "The Duration of the Action," shown in the third step "Heat the mug... for one minute"; and 3) "The Degree of the Action," shown in the final step "Stir... until well combined". Each step includes precise timestamps, such as 0:17 to 0:31 for the first segment.}
    \label{fig:Combined_Dataset_Visualization}
\end{figure*}

\subsection{Dataset Analysis} \label{subsec:Dataset Analysis}
Using the proposed pipeline, we construct the large-scale \textbf{DenseStep2M} dataset, comprising around 99k instructional videos drawn from HowTo100M, totaling over 7200 hours of footage with an average duration of 262 seconds per video. Each video is densely annotated with an average of 19 timestamped steps, averaging 10.4 words in length. Figure~\ref{fig:Dataset_Analysis_sub} illustrates the distribution of task categories, 
dominated by Food and Entertaining (59.1\%), followed by Hobbies and Crafts (17.4\%) and Home and Garden (11.4\%), along with histograms of step counts per video and word counts per step.

\vspace{3pt}\noindent \textbf{Comparison with Existing Datasets.}
Table~\ref{tab:Dataset Statistics.} compares DenseStep2M against several instructional video datasets, highlighting four key advantages.

\begin{itemize}[leftmargin=1em, itemindent=0pt]
    \item \textbf{High Step Density.} DenseStep2M contains an average of \textbf{19 steps per video}, nearly double the density of the next most detailed dataset (HowToStep~\cite{li2024multi}: 10.6 steps/video), enabling significantly finer-grained instructional analysis.

    \item \textbf{Rich, Multimodal-Grounded Descriptions.} As shown in Figure~\ref{fig:Compare_Dataset_sub}, step annotations incorporate \textbf{multimodal details} including purpose (``to ensure they are not overbaked''), precise durations (``cook for 2 minutes''), and ingredient quantities (``2 tablespoons of cornflour''), yielding richer and more actionable guidance than existing datasets.

    \item \textbf{Abstract and Context-Aware Annotations.} DenseStep2M captures \textbf{inferred, intent-driven steps} that go beyond observable content. For example, the annotation ``showcase a cat mug that is essential for the recipe''~(Figure~\ref{fig:Compare_Dataset_sub}) reflects the instructional intent of the video rather than its visual surface---an insight absent from other datasets. To our knowledge, DenseStep2M is the only dataset to include such context-dependent steps, supporting more advanced reasoning over instructional videos.

    \item \textbf{Rigorous Video Selection.} Through joint \textbf{visual-textual filtering}~(Sec.~\ref{subsec:Video Filtering and Segmentation}), only videos with at least 75\% instructional content are retained, ensuring high data relevance. In contrast, prior 
datasets rely solely on text-based filtering or lack systematic quality control altogether.
\end{itemize}

\begin{table}[h]
\centering
\small
\caption{\textbf{Comparison of DenseStep2M and other video datasets with instructional step annotations.}
}
\label{tab:Dataset Statistics.}
\begin{tabular}{lcccc}
\toprule
Dataset & Domain & \makecell[c]{\# Videos / \\ \# Steps} & \makecell[c]{\# Words \\per Step} & \makecell[c]{\# Steps \\per Video} \\ \midrule
CrossTask~\cite{zhukov2019cross}      & Open    & 4.7K / 21K  & 2.4           & 4.5           \\
YouCook2~\cite{zhou2018towards}       & Cooking & 2K / 14K    & 8.8           & 7.0           \\
COIN~\cite{tang2019coin}              & Open    & 11.8K / 46K & 4.8           & 3.9           \\
HT-Step~\cite{afouras2023ht}          & Cooking & 1.2K / 2.4K & -             & 2.0           \\
HowToStep~\cite{li2024multi}          & Cooking & 312K / 3.3M & 8.0           & 10.6          \\
ShowHowTo~\cite{souvcek2024showhowto} & Open    & 578K / 4.5M & 11.4          & 7.7           \\
HIREST~\cite{zala2023hierarchical}    & Open    & 3.4K / 8.6K & 4.4           & 2.5           \\ \midrule
DenseStep2M                         & Open    & 99.2K / 1.9M  & \textbf{10.5} & \textbf{19.0} \\ \bottomrule
\end{tabular}
\end{table}

\section{Downstream Applications} \label{sec:Downstream Applications}

To validate the utility and quality of DenseStep2M, we fine-tune state-of-the-art vision-language models and evaluate them on \textbf{dense video captioning}, \textbf{procedural step grounding}, and \textbf{cross-domain retrieval}, summarized in Table~\ref{tab:dataset_comparison}.

\subsection{Dense Video Captioning}
\textbf{Task Definition.} This task requires simultaneously localizing procedural events and generating timestamped natural-language descriptions to summarize long-form instructional videos, capturing both actions and specific states ({\em e.g.}, ``whisk until smooth'').

\vspace{3pt}\noindent \textbf{Benchmarks and Metrics.} We evaluate on \textbf{YouCook2} and our curated \textbf{DenseCaption100} (100 videos; 11.0 human-annotated steps/video). In addition to standard metrics (CIDEr~\cite{vedantam2015cider}, MET\-EOR~\cite{banerjee2005meteor}, SODA\_c~\cite{fujita2020soda}), we introduce the \textbf{R1-Score} via DeepSeek-R1 to measure \textbf{R1-Precision}  (measuring hallucination rate), \textbf{R1-Recall} (measuring procedural completeness), and \textbf{R1-mIoU} (temporal accuracy). Detailed definitions are in Appendix.

\vspace{3pt}\noindent \textbf{Implementation Details.} We fine-tune \textbf{Qwen2.5-VL-3B} using LoRA~\cite{hu2022lora} on both language and vision encoders. The model is trained for 1 epoch with a batch size of 8 videos across 4 NPUs. During training, we sample 50 frames ($224\times224$) and provide ASR transcripts as additional context to assist in generating dense narratives. For inference, the frame count is increased to 100 ($436\times250$) to capture longer temporal dependencies.

\subsection{Procedural Step Grounding}
\textbf{Task Definition.} Grounding aligns predefined textual instructions to temporal video segments, evaluating the model's ability to map linguistic cues to visual actions within complex sequences.

\vspace{3pt}\noindent \textbf{Benchmarks and Metrics.} Performance is assessed on \textbf{DenseCaption100} using \textbf{IoU} and overall \textbf{mIoU}. 

\vspace{3pt}\noindent \textbf{Implementation Details.} The architecture and LoRA setup remain consistent with captioning, but the input is modified to video frames paired with target steps as queries. We fine-tune the model using the VLLM-annotated steps from \textbf{DenseStep2M}, while evaluation is conducted on the human-annotated, canonical sequences from \textbf{DenseCaption100}.

\begin{table}[t]
\centering
\caption{Summary of evaluation benchmarks and metrics. We list evaluation metrics, average clip/video length (\#Len), and the number of test samples (\#Num).}
\small
\label{tab:dataset_comparison}
\setlength{\tabcolsep}{1.5pt} 
\begin{tabular}{lccc}
\toprule
Dataset & Evaluation Metrics & \#Len(s) & \#Num \\ \midrule
 \rowcolor{gray!10}\multicolumn{4}{c}{\textit{Video Captioning}} \\
YouCook2~\cite{zhou2018towards} & SODA\_c, CIDER, METEOR & 212.4 & 436 \\
DenseCaption100 & R1-score & 247.5 & 100 \\ \midrule
 \rowcolor{gray!10}\multicolumn{4}{c}{\textit{Step Grounding}} \\
DenseCaption100 & IoU@0.3, mIoU & 22.5 & 1100 \\ \midrule \midrule
\rowcolor{gray!10}\multicolumn{4}{c}{\textit{Egocentric benchmark}} \\
EgoMCQ~\cite{lin2022egocentric} & inter-/intra-video acc. & 34.2 & 39751 \\ \midrule
\rowcolor{gray!10}\multicolumn{4}{c}{\textit{Exocentric benchmark}} \\
YouCook2~\cite{zhou2018towards} & Video2Text R@1, R@5 & 19.7 & 3350 \\
MSRVTT~\cite{xu2016msr} & Text2Video/Video2Text R@1 & 14.8 & 1000 \\
DiDeMo~\cite{anne2017localizing} & Text2Video/Video2Text R@1 & 25\textasciitilde30 & 1004 \\ \midrule
\rowcolor{gray!10}\multicolumn{4}{c}{\textit{Ego-Exo benchmark}} \\
CharadesEgo~\cite{sigurdsson2018charades} & Ego2Exo/Exo2Ego R@1 & 22.7 & 145 \\
EgoExoBench~\cite{he2025egoexobench} & Task/Action/Object Relation acc. & - & 2240 \\ \bottomrule
\end{tabular}
\end{table}

\subsection{Cross-Modal Retrieval} 
We investigate the model's zero-shot generalization and perspective-invariant representations across \textbf{Video-to-Text (V2T), Text-to-Video (T2V), and Video-to-Video (V2V)} retrieval tasks. This evaluation spans egocentric, exocentric, and mixed-perspective (ego-exo) domains to assess the dataset's capacity to facilitate cross-modal action understanding.

\vspace{3pt}\noindent \textbf{Benchmarks and Metrics.}
The evaluation utilizes a diverse array of benchmarks: EgoMCQ for egocentric understanding; You\-Cook2, MSRVTT, and DiDeMo for exocentric retrieval; and CharadesEgo and EgoExoBench for cross-perspective alignment. Metrics include Recall@k and task-specific accuracy.

\vspace{3pt}\noindent \textbf{Implementation Details.}
Retrieval experiments are conducted on $8 \times$ A100 (80GB) GPUs. We evaluate two variants:

\begin{itemize}[leftmargin=1em, itemindent=0pt]
\item \textbf{Qwen3-VL-Embedding-8B:} We apply LoRA ($r=32, \alpha=32$) to fine-tune both the vision and language encoders. We sample 8 frames per clip at $256 \times 256$ resolution. Hard negatives are mined via global cosine similarity, retrieving 7 candidates per query with a $\ge 0.05$ similarity margin.
\item \textbf{Qwen3-VL-8B-Instruct:} We fine-tune only the language model using high-rank LoRA ($r=128, \alpha=256$). Each video is represented by 8 frames. We use a distributed similarity batch sampler where 80\% of negatives are hard samples with text similarity scores $>0.5$.
\end{itemize}

Both models are optimized via a \textbf{Symmetric Masked InfoNCE} loss~\cite{oord2018representation}. The total loss $\mathcal{L}$ is defined as the average of Video-to-Text ($\mathcal{L}_{v2t}$) and Text-to-Video ($\mathcal{L}_{t2v}$) components. For batch size $B$, the Video-to-Text component $\mathcal{L}_{v2t}$ is defined as:
\begin{equation}
\mathcal{L}_{v2t} = -\frac{1}{B} \sum_{i=1}^{B} \log \frac{\exp(s_{i,i}/\tau)}{\sum_{j=1}^{B} \mathbbm{1}[\text{mask}_{i,j}] \cdot \exp(s_{i,j}/\tau)}
\end{equation}
where $s_{i,j}$ is the cosine similarity, and $\tau$ is set to 0.03 (Instruct) or 0.05 (Embedding). To handle false negatives, pairs with text similarity $\eta > 0.85$ are masked ($\text{mask}_{i,j}=0$). Training lasts 1 epoch with global batch sizes of 768 (Instruct) and 256 (Embedding).

\section{Experiments and Evaluations} \label{sec:Experiments}

We evaluate both the proposed \textbf{annotation pipeline} and the resulting 
\textbf{DenseStep2M dataset} along two axes: 
(i)~\textbf{annotation quality}: does the automated pipeline produce semantically accurate and temporally precise labels relative to human ground truth? 
and (ii)~\textbf{downstream utility}: does training on DenseStep2M consistently improve model performance and generalization in long-form video understanding across diverse viewpoints? 

Specifically, we first detail the implementation of our pipeline (Sec.~\ref{subsec:Implementation Details}), followed by a quantitative assessment of annotation quality and a comprehensive ablation study (Sec.~\ref{subsec:Evaluation of Pipeline Quality}). We then demonstrate the utility of DenseStep2M across several downstream tasks, including dense video captioning, procedural grounding, and cross-modal retrieval (Sec.~\ref{subsec:Downstream Task Evaluation}).

\subsection{Implementation Details of the Pipeline} \label{subsec:Implementation Details}

The pipeline sequentially employs 3 frontier models to process the HowTo100M source data. For the \textbf{segmentation stage}, Qwen2.5-72B~\cite{qwen2.5} is deployed on 4 NPUs with a request concurrency of 100. In the \textbf{generation stage}, Qwen2.5-VL-72B~\cite{bai2025qwen2} samples frames at 1 FPS (capped at 80 frames per shot), distributed across 16 NPUs with a concurrency of 10, achieving a daily throughput of approximately 6,000 video segments (avg. duration: 50s). Finally, the \textbf{refinement stage} utilizes DeepSeek-R1-671B~\cite{guo2025deepseek} on 16 NPUs with a concurrency of 16, reaching a decoding throughput of 275 tokens/s.

\subsection{Evaluation of Pipeline Quality} \label{subsec:Evaluation of Pipeline Quality}
In this section, we evaluate the quality of the DenseStep2M pipeline on the public benchmark \textbf{YouCook2} and our curated \textbf{DenseCaption100}, followed by an ablation study to isolate the contributions of our key design choices.

\begin{table}[t]
\centering
\caption{\textbf{Comparison with state-of-the-art on YouCook2 (val).}}
\label{tab:Comparison with State-of-the-Art}
\begin{tabular}{lcccc}
\toprule 
Method & Pretrain & SODA\_c $\uparrow$ & CIDEr $\uparrow$ & METEOR $\uparrow$ \\
\midrule
Vid2Seq~\cite{yang2023vid2seq} & \checkmark & 7.9 & 47.1 & 9.3 \\
HiCM\textsuperscript{2}~\cite{kim2025hicm2} & \checkmark & 10.7 & 71.8 & 12.8 \\
\midrule
E2ESG~\cite{zhu2022end} & \texttimes & — & 25.0 & 3.5 \\
Vid2Seq~\cite{yang2023vid2seq} & \texttimes & 4.0 & 18.0 & 4.6 \\
PDVC~\cite{wang2021end} & \texttimes & 4.9 & 28.9 & 5.7 \\
CM\textsuperscript{2}~\cite{kim2024you} & \texttimes & \underline{5.3} & \underline{31.7} & \underline{6.1} \\
Ours & \texttimes & \textbf{5.6} & \textbf{33.6} & \textbf{9.4} \\
\bottomrule
\end{tabular}
\end{table}

\begin{table}[!t]
\centering
\caption{\textbf{Architecture comparison on DenseCaption100.} \textit{Time}: steps timestamped by VLLM; \textit{R1}: outputs refined by DeepSeek-R1~\cite{guo2025deepseek}; \textit{mIoU}: R1-mIoU metric. VLLM sampling is 1 FPS (max 80/300 frames for shot/raw video). \textsuperscript{*} denotes our default setting.}
\label{tab:Compare architectures on DenseCaption100}
\small
\setlength{\tabcolsep}{2pt}
\begin{tabular}{llcccccccc}
\toprule
No.  & \multicolumn{1}{c}{VLLM} & Visual Input & Size & Time & R1       & P $\uparrow$   & R $\uparrow$ & F1 $\uparrow$  & mIoU $\uparrow$ \\ \midrule
1. & Qwen2.5 & raw video  & 72B & \checkmark & \texttimes & \textbf{73.05} & 76.56          & \underline{74.76} & 25.45 \\ \midrule
2. & Qwen2   & video shot & 72B & \checkmark & \checkmark & 62.19 & 78.77          & 69.50 & \underline{42.74} \\
3. & Qwen2.5 & video shot & 7B  & \checkmark & \checkmark & 57.89 & 80.47          & 67.34 & 42.58 \\ \midrule
4. & Qwen2.5 & video shot & 72B & \checkmark & \texttimes & 57.39 & \textbf{86.57} & 69.03 & 32.87 \\
5. & Qwen2.5 & video shot & 72B & \texttimes & \checkmark & 59.75 & 82.63          & 69.35 & 41.16 \\ \midrule
6.\textsuperscript{*} & Qwen2.5                  & video shot   & 72B  & \checkmark      & \checkmark & \underline{67.79} & \underline{85.97}        & \textbf{75.81} & \textbf{42.90}     \\ \bottomrule
\end{tabular}
% }
\end{table}

\begin{table}[!h]
\centering
\caption{\textbf{Utility of DenseStep2M (DS) for fine-tuning VLLMs on DenseCaption100.} \textit{FT Data}: fine-tuning source; \textit{ASR}: ASR-only    ; \textit{mIoU}: R1-mIoU. Bold indicates the best performance using our proposed high-quality dataset.}
\label{tab:Performance Improvement of VLLMs with DenseStep2M Fine-Tuning.}
\small
\setlength{\tabcolsep}{1.5pt}
\begin{tabular}{llcccccc}
\toprule
\multirow{2}{*}{Method} & \multirow{2}{*}{Size} & \multirow{2}{*}{FT Data} & \multicolumn{3}{c}{Dense Video Captioning} & \multicolumn{2}{c}{Step Grounding} \\ \cmidrule(lr){4-6} \cmidrule(lr){7-8} 
 & & & P $\uparrow$ & R $\uparrow$ & R1-mIoU $\uparrow$ & IoU@0.3 $\uparrow$ & mIoU $\uparrow$ \\ \midrule
Qwen2.5-VL & 3B & \texttimes & 51.16 & 62.55 & 28.04 & 4.64 & 3.13 \\
Qwen2.5-VL & 3B & DS-ASR & 49.28 & 68.23 & 37.49 & 40.52 & 26.04 \\
\textbf{Qwen2.5-VL} & \textbf{3B} & \textbf{DS} & \textbf{55.29} & \textbf{73.55} & \textbf{41.70} & \textbf{50.14} & \textbf{34.65} \\ \bottomrule
\end{tabular}
\end{table}

\begin{table*}[!ht]
\centering
\caption{\textbf{Cross-Domain Video Retrieval Results} across Ego-centric, Exo-centric, and Ego-Exo benchmarks. DS2M refers to the DenseStep2M dataset. Parentheses indicate absolute gains/losses over the respective baseline.}
\label{tab:Cross-Domain Retrieval Results}
\setlength{\tabcolsep}{3pt}
\resizebox{\textwidth}{!}{
\begin{tabular}{lccccccccccccc}
\toprule
\multirow{3}{*}{Method} & \multicolumn{2}{c}{Ego-centric Benchmark} & \multicolumn{6}{c}{Exo-centric Benchmark} & \multicolumn{5}{c}{Ego-Exo Benchmark} \\
\cmidrule(lr){2-3} \cmidrule(lr){4-9} \cmidrule(lr){10-14}
 & \multicolumn{2}{c}{EgoMCQ} & \multicolumn{2}{c}{YouCook2} & \multicolumn{2}{c}{MSRVTT} & \multicolumn{2}{c}{DiDeMo} & \multicolumn{3}{c}{EgoExoBench} & \multicolumn{2}{c}{CharadesEgo} \\
\cmidrule(lr){2-3} \cmidrule(lr){4-5} \cmidrule(lr){6-7} \cmidrule(lr){8-9} \cmidrule(lr){10-12} \cmidrule(lr){13-14}
 & Inter Acc. & Intra Acc. & R@1 & R@5 & T2V & V2T & T2V & V2T & TR & AR & OR & Ego2Exo & Exo2Ego \\ \midrule
 \rowcolor{gray!10}\multicolumn{14}{c}{\textit{Comparison Models}} \\
% \textit{Comparison Models} & & & & & & & & & & & & & \\
GME-7B~\cite{zhang2024gme} & - & - & 9.1 & - & 31.8 & - & 26.4 & - & - & - & - & - & - \\
LamRA-Qwen2.5~\cite{liu2025lamra} & - & - & 7.5 & - & 25.0 & - & 22.8 & - & - & - & - & - & - \\
VLM2Vec-V2.0~\cite{meng2025vlm2vec} & - & - & 10.6 & - & 28.3 & - & 30.4 & - & - & - & - & - & - \\
InternVL3-8B~\cite{zhu2025internvl3} & - & - & - & - & - & - & - & - & 36.8 & 30.9 & 37.0 & - & - \\
LLaVA-Video-7B~\cite{zhang2024llava} & - & - & - & - & - & - & - & - & 33.9 & 29.1 & 35.4 & - & - \\ \midrule
 \rowcolor{gray!10}\multicolumn{14}{c}{\textit{Embedding Models}} \\
% \textit{Embedding Models} & & & & & & & & & & & & & \\
Qwen3-VL-emb-8B~\cite{li2026qwen3} & 84.5 & 38.2 & 31.0 & \textbf{58.7} & \textbf{55.5} & 52.0 & \textbf{60.1} & \textbf{56.5} & 54.0 & \textbf{41.2} & \textbf{54.7} & \textbf{80.7} & \textbf{75.2} \\
\textbf{+ DS2M} & \textbf{84.8} \scriptsize{(+0.3)} & 38.2 \scriptsize{(0)} & \textbf{36.4} \scriptsize{(+5.4)} & 58.0 \scriptsize{(-0.7)} & 51.5 \scriptsize{(-4.0)} & \textbf{52.6} \scriptsize{(+0.6)} & 59.1 \scriptsize{(-1.0)} & 55.7 \scriptsize{(-0.8)} & \textbf{55.1} \scriptsize{(+1.1)} & 40.3 \scriptsize{(-0.9)} & 53.2 \scriptsize{(-1.5)} & 78.6 \scriptsize{(-2.1)} & 70.3 \scriptsize{(-4.9)} \\ \midrule
 \rowcolor{gray!10}\multicolumn{14}{c}{\textit{Instruct Models}} \\
% \textit{Instruct Models} & & & & & & & & & & & & & \\
Qwen3-VL-8B~\cite{bai2025qwen3} & 67.6 & 32.2 & 6.4 & 18.3 & 30.5 & 30.3 & 23.1 & 18.3 & 37.2 & 34.7 & 46.4 & 29.7 & 24.1 \\
\textbf{+ DS2M} & 80.7 \scriptsize{(+12.4)} & 36.7 \scriptsize{(+4.5)} & \textbf{35.4} \scriptsize{(+29.0)} & \underline{63.0} \scriptsize{(+44.7)} & 30.5 \scriptsize{(0)} & \textbf{40.0} \scriptsize{(+9.7)} & 35.3 \scriptsize{(+12.2)} & \textbf{38.5} \scriptsize{(+20.2)} & \textbf{40.9} \scriptsize{(+3.7)} & 38.3 \scriptsize{(+3.6)} & 48.8 \scriptsize{(+2.4)} & \textbf{66.9} \scriptsize{(+37.2)} & \textbf{48.3} \scriptsize{(+24.2)} \\
+ Ego4D~\cite{grauman2022ego4d} & \underline{92.9} \scriptsize{(+25.3)} & \textbf{49.3} \scriptsize{(+17.1)} & 15.9 \scriptsize{(+9.5)} & 36.0 \scriptsize{(+17.7)} & 29.0 \scriptsize{(-1.5)} & 30.5 \scriptsize{(+0.2)} & \textbf{37.2} \scriptsize{(+14.1)} & 34.9 \scriptsize{(+16.6)} & 36.6 \scriptsize{(-0.6)} & \textbf{43.6} \scriptsize{(+8.9)} & \textbf{53.1} \scriptsize{(+6.7)} & 55.2 \scriptsize{(+25.5)} & 25.5 \scriptsize{(+1.4)} \\
\textbf{+ Ego4D + DS2M} & \textbf{93.0} \scriptsize{(+25.4)} & \underline{48.3} \scriptsize{(+16.1)} & \underline{34.7} \scriptsize{(+28.3)} & \textbf{63.1} \scriptsize{(+44.8)} & \textbf{32.1} \scriptsize{(+1.6)} & \underline{37.1} \scriptsize{(+6.8)} & 33.0 \scriptsize{(+9.9)} & \underline{36.4} \scriptsize{(+18.1)} & 37.0 \scriptsize{(-0.2)} & \underline{41.4} \scriptsize{(+6.7)} & \underline{51.8} \scriptsize{(+5.4)} & \underline{62.1} \scriptsize{(+32.4)} & \underline{46.2} \scriptsize{(+22.1)} \\ \bottomrule
\end{tabular}
}
\end{table*}

\vspace{3pt}\noindent \textbf{Benchmark Verification.}
% \subsubsection{Benchmark Verification}
We first verify the pipeline's effectiveness on the \textbf{YouCook2} validation set. As presented in Table~~\ref{tab:Comparison with State-of-the-Art}, our training-free method achieves a METEOR score of 9.4, surpassing the previous state-of-the-art without pretraining (6.1 by CM$^2$~\cite{kim2024you}) by 3.3 points (+54.1\%), indicating substantially improved semantic alignment with human captions. Similarly, our method attains a CIDEr score of 33.6 and a $SODA_c$ score of 5.6, exceeding the SOTA without pretraining (31.7 and 5.3 by CM$^2$~\cite{kim2024you}) by 1.9 points (+6.0\%) and 0.3 points (+5.7\%), respectively.

On our newly curated \textbf{DenseCaption100} benchmark (Table~\ref{tab:Compare architectures on DenseCaption100}), the full pipeline (Row 6) significantly outperforms the raw VLLM baseline (Row 1) in R1-Recall (86.0), F1 (75.8), and mIoU (42.9). While Row 1 exhibits higher Precision (73.1), suggesting the baseline is more conservative, our approach captures a much broader and more temporally accurate set of instructional steps.

\vspace{3pt}\noindent \textbf{Ablation Analysis.}
% \subsubsection{Ablation Analysis}
To further understand the factors driving these performance gains, we conduct an ablation study on the DenseCaption100 benchmark (Table~\ref{tab:Compare architectures on DenseCaption100}).

\begin{itemize}[leftmargin=1em, itemindent=0pt]
\item \textbf{VLLM Capacity.} Upgrading from Qwen2.5-VL-7B or Qwen2-VL-72B to Qwen2.5-VL-72B (Row 6 vs. 2, 3) yields substantial improvements across all metrics. This confirms that high-capacity VLLMs are critical for recognizing the fine-grained procedural details required for dense annotation.

\item \textbf{Refinement Stage.} The integration of DeepSeek-R1 serves as a critical "global auditor" for the pipeline. Comparing Row 6 (Full Pipeline) with Row 4 (VLLM-only with timestamps), we observe a substantial improvement in both Precision (from 57.4\% to 67.8\%) and mIoU (from 32.9\% to 42.9\%). This indicates that while VLLMs are adept at visual recognition, they often produce "fragmented hallucinations" or redundant steps when faced with complex multi-modal cues. DeepSeek-R1 rectifies these artifacts by performing cross-segment logical verification, merging semantically overlapping actions, and filtering out steps that lack causal consistency with the overall video narrative.

\item \textbf{Initial Scaffolding.} Our results also highlight a vital "scaffolding" effect. Comparing Row 6 with Row 5 (where no initial VLLM timestamps are provided), the inclusion of noisy visual timestamps from the VLLM leads to a noticeable boost in mIoU. This suggests that these initial visual anchors, even if imprecise, provide a spatial-temporal context that text-only reasoning cannot fully recover from ASR transcripts alone. The synergy between the VLLM’s "perceptual scaffold" and the Reasoning LLM’s "logical rectification" is the key driver behind the full pipeline's excellent performance on the DenseCaption100 benchmark.

\end{itemize}

\subsection{Performance in Downstream Tasks} \label{subsec:Downstream Task Evaluation}
To evaluate the extrinsic utility of DenseStep2M, we conduct extensive experiments on three core tasks as described in Sec.~\ref{sec:Downstream Applications}. 

\vspace{3pt}\noindent \textbf{Captioning and Grounding.}
We evaluate the impact of fine-tuning \textbf{Qwen2.5-VL-3B} on our dataset. To isolate the effect of dataset quality from the benefits of task-specific fine-tuning, we introduce a control dataset, \textbf{DenseStep2M-ASROnly}. This baseline utilizes the same source videos but generates steps via a vision-agnostic approach, prompting a text-only Qwen2.5-72B-Instruct to hallucinate instructions based solely on ASR transcripts.

As shown in Table~\ref{tab:Performance Improvement of VLLMs with DenseStep2M Fine-Tuning.}, fine-tuning on our \textbf{DenseStep2M} yields substantial and consistent gains. For \textbf{Dense Video Captioning}, the model achieves superior semantic alignment and temporal accuracy, reaching an R1-mIoU of 41.7\%—a \textbf{+4.2\%} absolute improvement over the vision-agnostic counterpart. This is further supported by robust semantic coverage, with R1-Precision ($P$) and R1-Recall ($R$) increasing by \textbf{+6.0\%} and \textbf{+5.3\%}, respectively. The boost is even more pronounced in \textbf{Step Grounding}, where mIoU reaches 34.7\%, outperforming the baseline by \textbf{+8.6\%}. Compared to ASR-only alternatives, our richly annotated data significantly enhances the model's ability to ground complex procedural events.

\vspace{3pt}\noindent \textbf{Cross-Modal Retrieval.}
We investigate the scalability and generalization of DenseStep2M
via cross-modal retrieval across egocentric, exocentric, and mixed-perspective domains (Table~\ref{tab:Cross-Domain Retrieval Results}).

Fine-tuning \textbf{Qwen3-VL-embedding-8B} on the full 1.9M-step dataset significantly boosts expertise in instructional contexts, as evidenced by a 3\textbf{6.4\% R@1 (+5.4\%)} on YouCook2. While the model gains a profound understanding of procedural logic, we observe a slight performance trade-off on general-purpose benchmarks like MSRVTT, likely due to domain-specific over-specialization. We posit that incorporating a more diverse dataset mixture during fine-tuning could mitigate this effect and further regularize the embedding space.

For the \textbf{Qwen3-VL-8B-Instruct} model, fine-tuning on DenseStep2M proves transformative, yielding substantial performance leaps across all categories. Notably, the dataset’s rich procedural logic bolsters exocentric performance, with YouCook2 R@1 soaring by \textbf{+29.0\%}—markedly outperforming the model fine-tuned on Ego4D alone. Performance on ego-exo benchmarks remains highly competitive, demonstrating that our granular, step-by-step descriptions provide a superior supervisory signal for mapping complex visual procedures to linguistic tokens.

A critical challenge in egocentric pre-training (e.g., on Ego4D) is the resulting "exocentric blind spot," where the model excels at first-person tasks but falters in third-person instructional scenarios. Our results show that the hybrid configuration (\textbf{Ego4D + DS2M}) yields the most robust performance, maintaining peak egocentric accuracy (93.0 on EgoMCQ Inter) while simultaneously eliminating the performance gap in exocentric benchmarks. For instance, adding DS2M to an Ego4D-trained model boosts MSRVTT V2T R@1 from 30.5\% to 37.1\%. This empirical evidence confirms that DenseStep2M provides a necessary exocentric counterbalance, facilitating the development of truly perspective-invariant representations.

\section{Conclusion}

In this paper, we present a novel, training-free pipeline for automatically annotating instructional videos with dense, timestamped procedural steps, resulting in the DenseStep2M dataset. Our experimental results highlight the pipeline’s state-of-the-art performance on YouCook2 and its substantial superiority over a leading VLLM on DenseCaption100, alongside notable performance gains in downstream tasks after fine-tuning on DenseStep2M. These advancements underscore the potential of our approach and dataset to enhance video understanding across diverse applications, from education to robotics, setting a foundation for future research in scalable video annotation and analysis.

\clearpage
\bibliographystyle{ACM-Reference-Format}
\bibliography{references}

\clearpage
\appendix

\section*{Appendix}
In the appendix, we include the following content: Dataset generation details (Appendix~\ref{sec:Dataset Generation Details}), Datasets and evaluation metrics (Appendix~\ref{sec:Datasets Evaluation Metrics}), Implementation details (Appendix~\ref{sec:Implementation Details}), Additional quantitative results (Appendix~\ref{sec:Additional Quantitative Results}), Additional qualitative results (Appendix~\ref{sec:Additional Qualitative Results}), Limitations and ethical concerns (Appendix~\ref{sec:Limitations and Ethical Concerns}) and Broader impacts (Appendix~\ref{sec:Broader Impacts}).

\section{Dataset Generation Details}\label{sec:Dataset Generation Details}

\begin{figure*}[th]
    \centering
    \begin{subfigure}[t]{0.46\textwidth}
        \includegraphics[width=\textwidth]{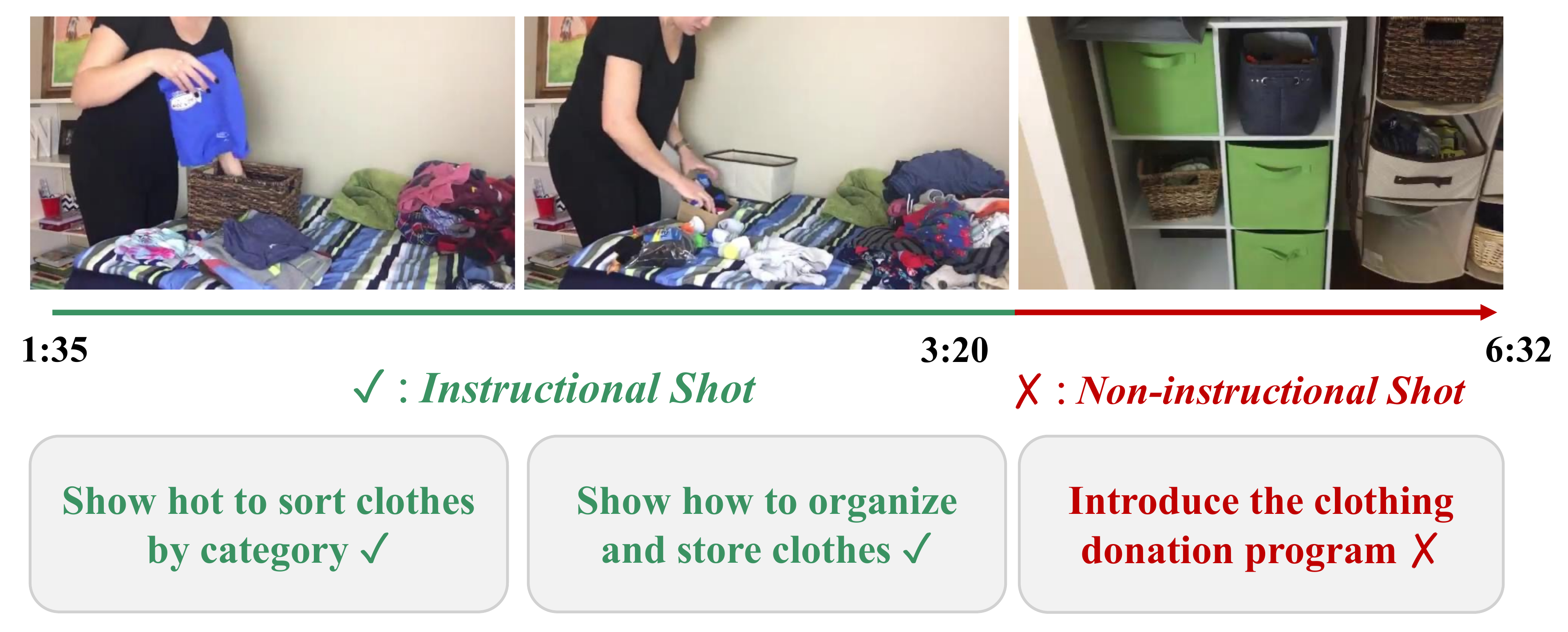}
        \caption{\textbf{A video dominated by non-instructional content}, such as personal anecdotes or unrelated dialogue, reducing its instructional value.}
        \label{fig:Non-instructional_Segment}
    \end{subfigure}
    \hfill  
    \begin{subfigure}[t]{0.5\textwidth}
        \includegraphics[width=\textwidth]{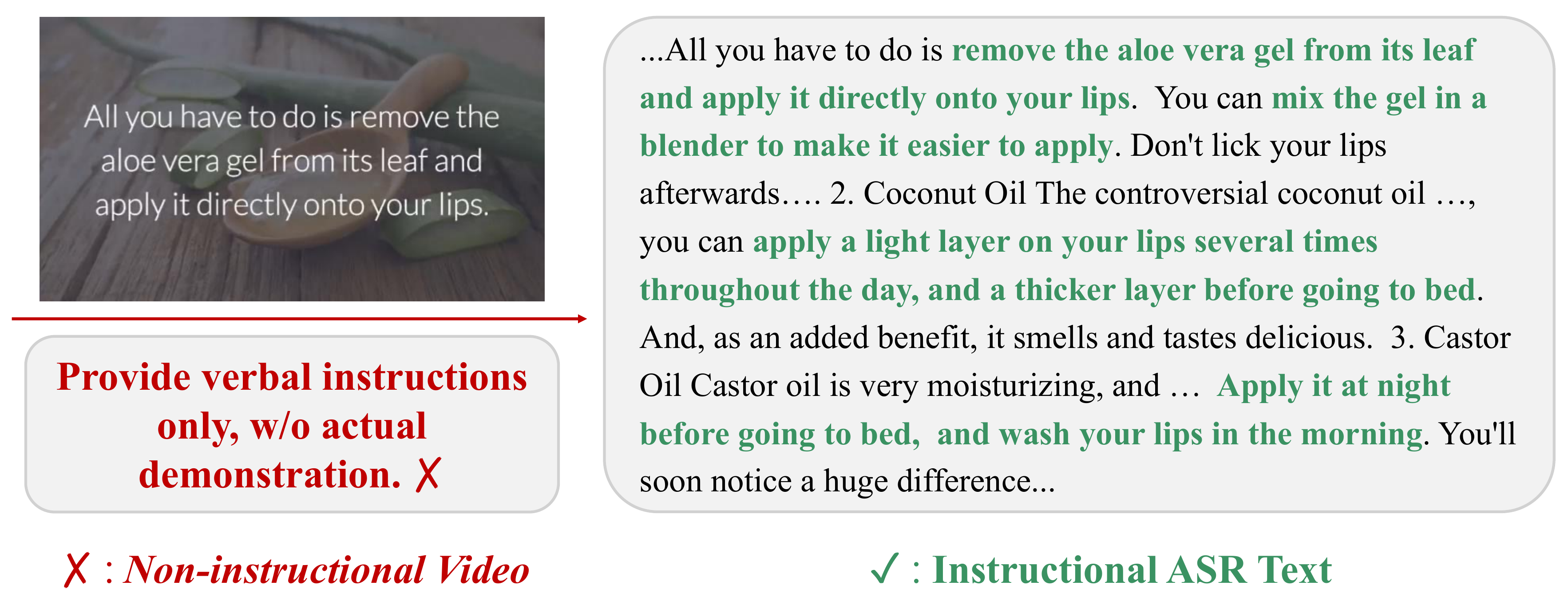}
        \caption{\textbf{A non-instructional video incorrectly paired with instructional ASR transcripts}, demonstrating the need for accurate cross-modal alignment.}
        \label{fig:Non-instructional_Video}
    \end{subfigure}
    \caption{\textbf{Examples of low-quality videos from the HowTo100M dataset}~\cite{miech2019howto100m}.  
    }  
    \label{fig:overall}
\end{figure*}

\begin{figure*}[th]
    \centering
    \includegraphics[width=0.8\textwidth]{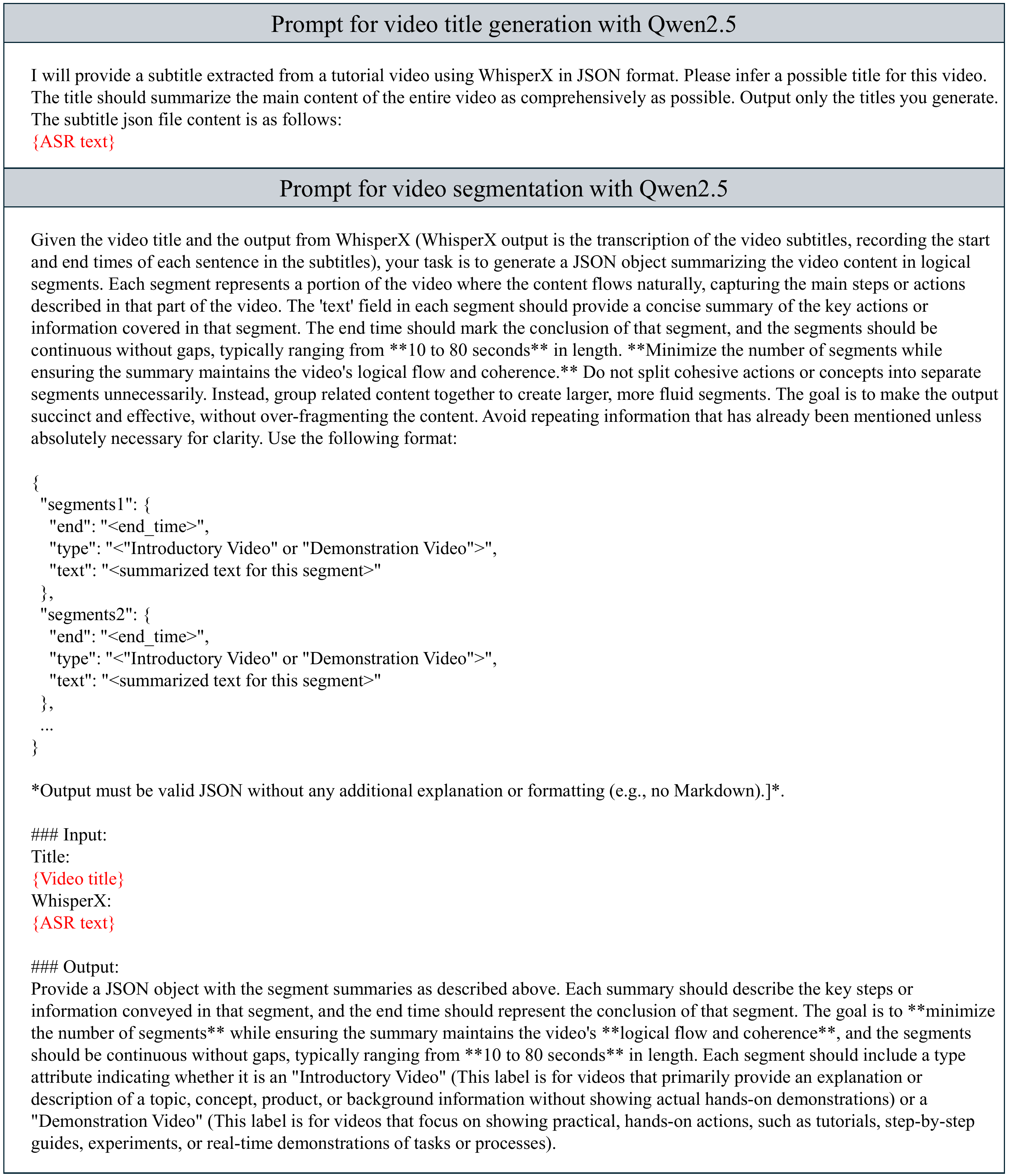}
    \caption{\textbf{Prompts for splitting the video and generating the title with Qwen2.5.} Relevant examples have been omitted for clarity, and the full prompts can be found in our code.}
    \label{fig:Stage1_details}
\end{figure*}

\begin{figure*}[th]
    \centering
    \includegraphics[width=0.8\textwidth]{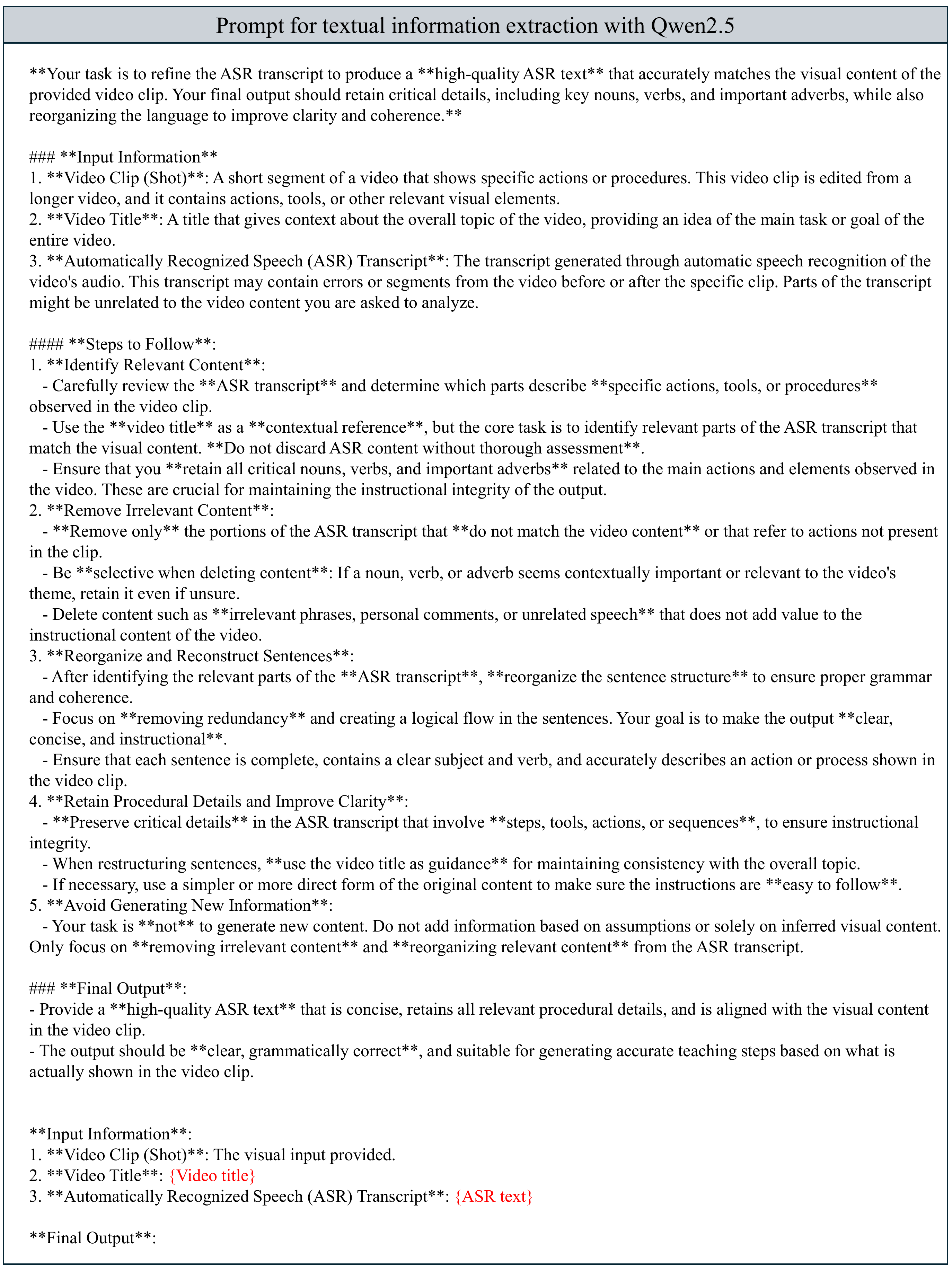}
    \caption{\textbf{Prompts for extracting textual information with Qwen2.5.} Relevant examples have been omitted for clarity, and the full prompts can be found in our code.}
    \label{fig:texual_info_details}
\end{figure*}

\begin{figure*}[th]
    \centering
    \includegraphics[width=0.8\textwidth]{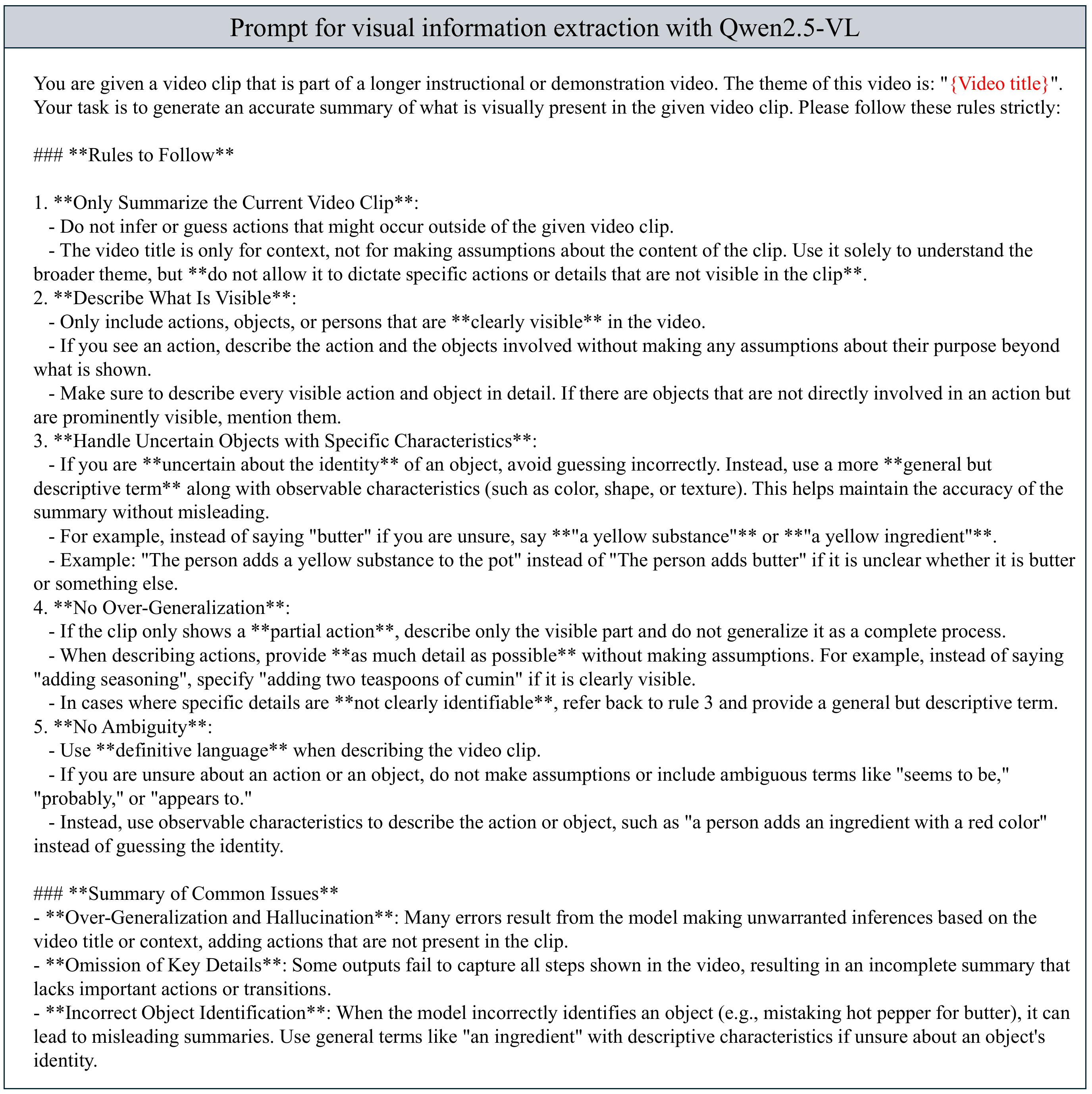}
    \caption{\textbf{Prompts for extracting visual information with Qwen2.5-VL.} Relevant examples have been omitted for clarity, and the full prompts can be found in our code.}
    \label{fig:visual_info_details}
\end{figure*}

\begin{figure*}[th]
    \centering
    \includegraphics[width=0.8\textwidth]{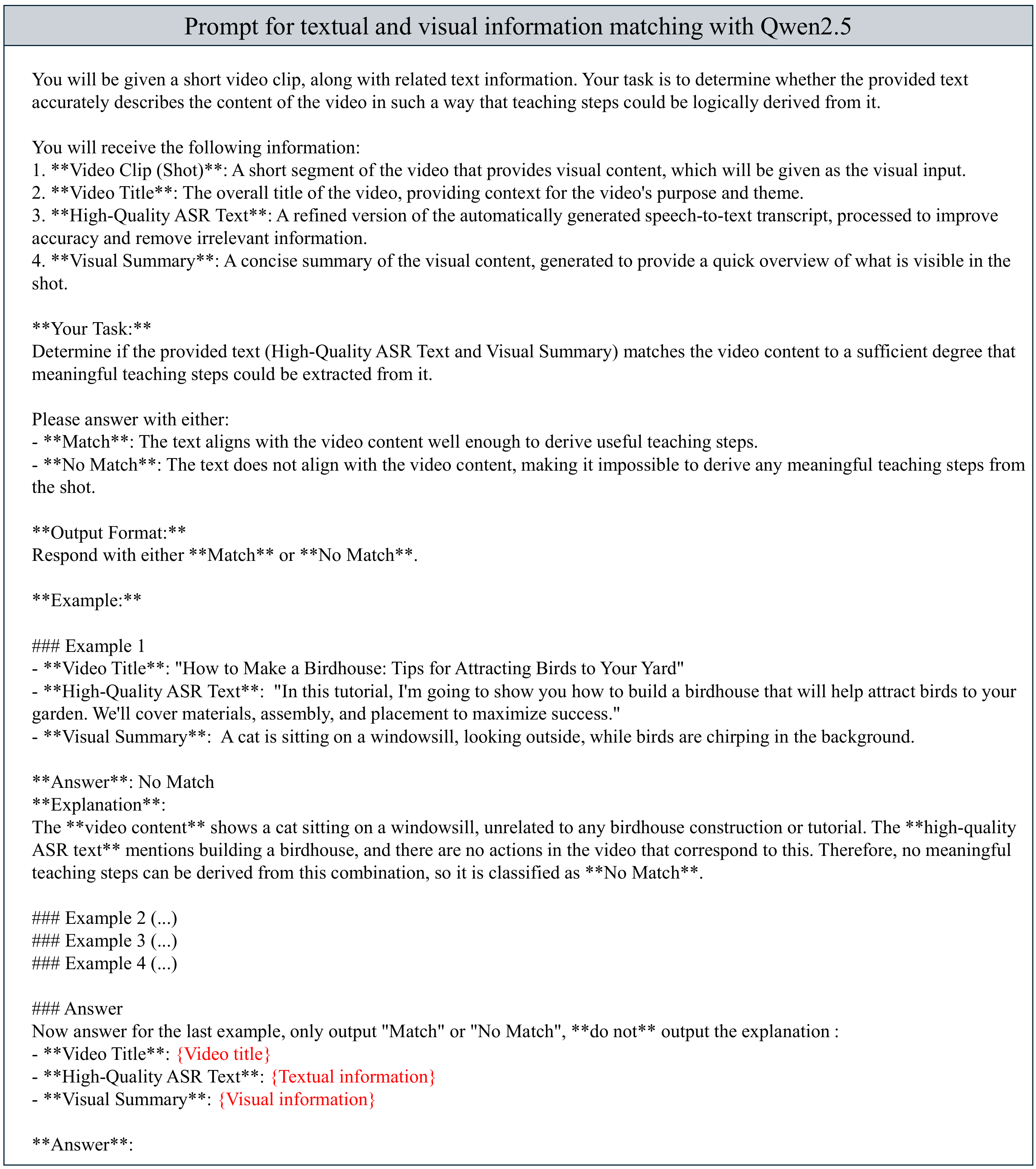}
    \caption{\textbf{Prompts for matching textual and visual information with Qwen2.5.} Relevant examples have been omitted for clarity, and the full prompts can be found in our code.}
    \label{fig:matching_details}
\end{figure*}

\textbf{Details on Video Filtering and Segmentation.} As described in Sec.~\ref{subsec:Video Filtering and Segmentation}
, the HowTo100M dataset~\cite{miech2019howto100m} offers a rich and varied video corpus, but its quality is highly inconsistent. Figure~\ref{fig:overall} illustrates typical low-quality examples, including wholly non-instructional content ({\em e.g.,} travel logs and personal anecdotes), videos containing lengthy non-instructional segments ({\em e.g.,} Figure~\ref{fig:Non-instructional_Segment}, where non-instructional footage overwhelms the tutorial content), and clips with mismatched modalities ({\em e.g.,} Figure~\ref{fig:Non-instructional_Video}, where ASR transcripts describe instructional steps despite non-instructional visuals). To isolate truly instructional material, we deployed a three-step multimodal filtering and segmentation pipeline on HowTo100M~\cite{miech2019howto100m} videos—first refining ASR outputs, then extracting and aligning textual and visual cues, and finally retaining only those videos with $\geq$75\% of their segments exhibiting matched instructional content. This process yielded 99.2 K (8.1\%) high-quality instructional videos while discarding 1.12 M (91.9\%) low-quality ones. The full prompts are detailed in Figure~\ref{fig:Stage1_details},~\ref{fig:texual_info_details},~\ref{fig:visual_info_details} and~\ref{fig:matching_details}.

\begin{figure*}[th]
    \centering
    \includegraphics[width=0.8\textwidth]{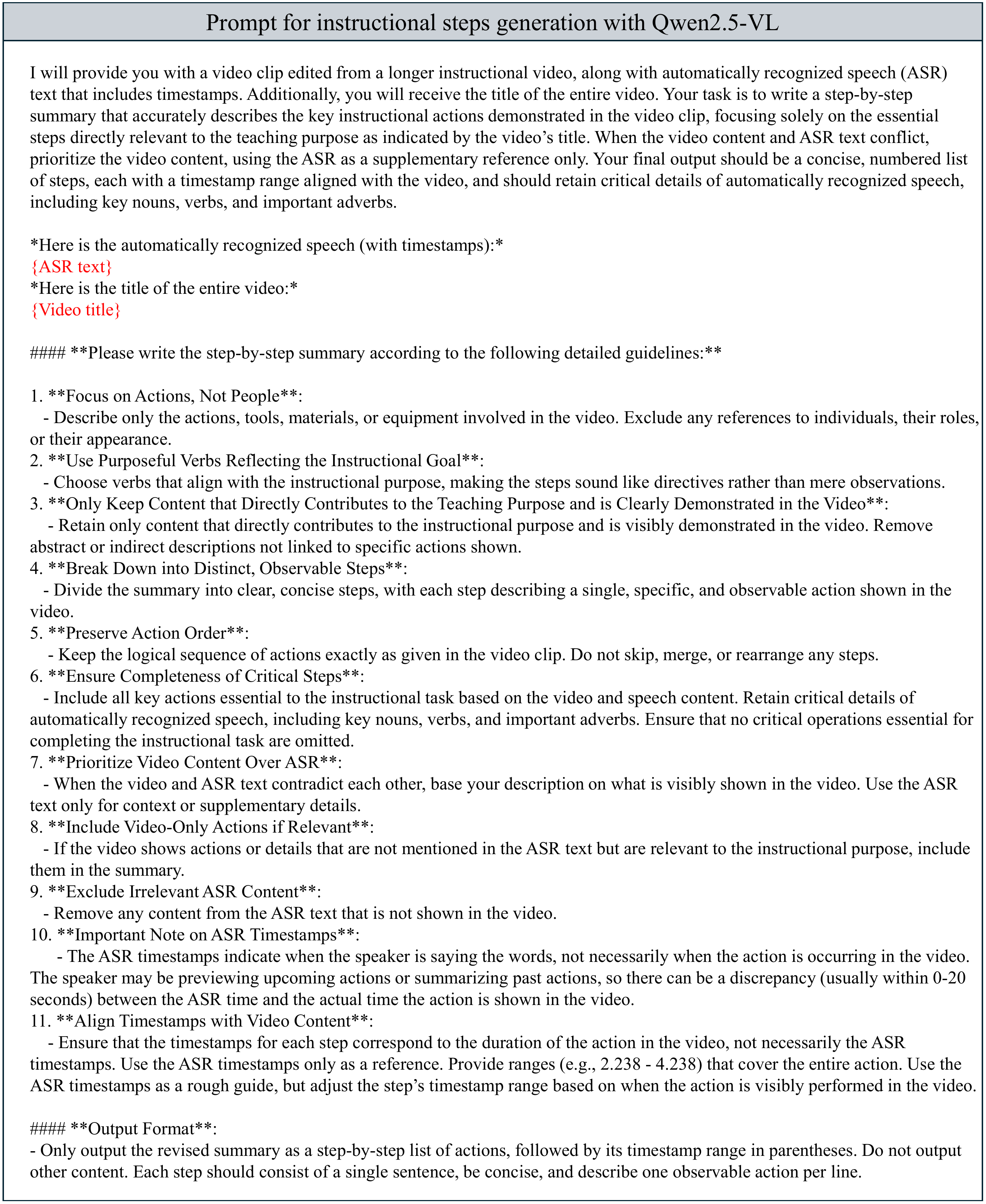}
    \caption{\textbf{Prompts for generating instructional steps with Qwen2.5-VL.} Relevant examples have been omitted for clarity, and the full prompts can be found in our code.}
    \label{fig:VLLM_Details}
\end{figure*}

\vspace{3pt}\noindent \textbf{Details on Instructional Steps Generation.} We aim to generate clear, timestamped instructional steps by first dividing each video into visually coherent shots—each represented by up to 80 frames sampled at 1 fps—and then leveraging Qwen2.5-VL-72B-Instruct~\cite{bai2025qwen2} to extract procedures from each segment (see Sec. ~\ref{subsec:Instructional Step Generation}
). For each shot, we supply its key frames, the video title, and the ASR transcript (which itself includes per-sentence timestamps) as inputs, enabling the model to output concise, natural-language steps linked to their start and end times. The full prompt used to guide the model is detailed in Figure~\ref{fig:VLLM_Details}. This process yields an initial, shot-level collection of semantically rich, temporally grounded instructions that can then be refined in subsequent stages.

\begin{figure*}[th]
    \centering
    \includegraphics[width=0.75\textwidth]{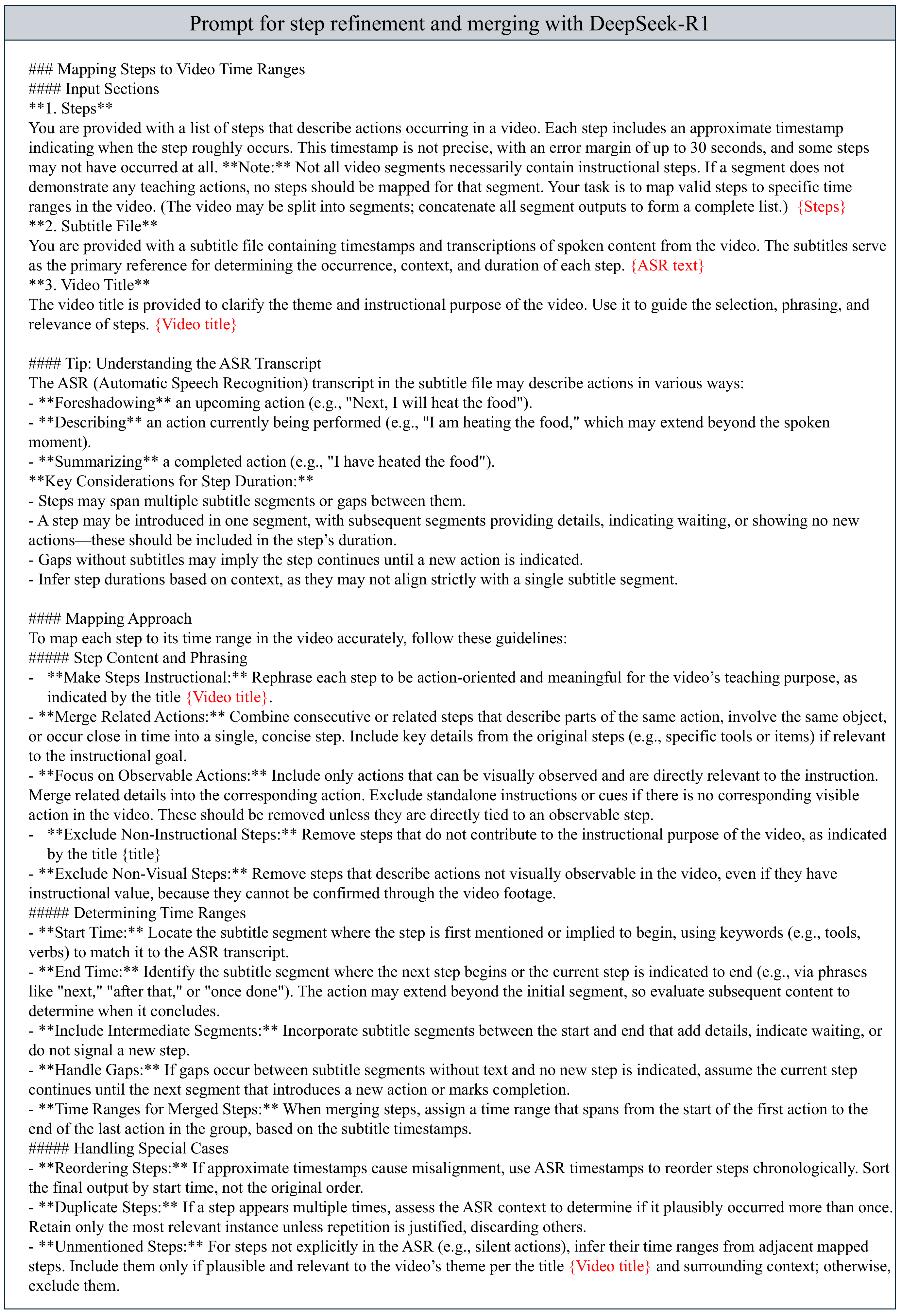}
    \caption{\textbf{Prompts for refining and merging steps with  DeepSeek-R1.} Relevant examples have been omitted for clarity, and the full prompts can be found in our code.}
    \label{fig:refinement_details}
\end{figure*}

\vspace{3pt}\noindent \textbf{Details on Refining and Merging Steps.} We observe that the raw shot-level steps produced by our vision–language model frequently suffer from out-of-order sequences, spurious or duplicated actions, and imprecise timing (see Sec. ~\ref{subsec:Step Refining and Merging}
). To correct these errors, we employ a reasoning LLM, DeepSeek-R1-671B~\cite{guo2025deepseek}, which ingests the initial step proposals alongside the full ASR transcript, subtitle timestamps, and the video title (full prompt specification in Figure~\ref{fig:refinement_details}).  The model first evaluates the likelihood of each candidate step at various timepoints, grounding them linguistically by matching keywords and temporal markers in the transcript to ensure that every proposed action is textually supported. Building on this grounding phase, DeepSeek-R1-671B~\cite{guo2025deepseek} refines the start and end timestamps by locating the first and last subtitle segments that correspond to each action—including foreshadowing or summary utterances—then filters out hallucinated or redundant proposals by discarding any step not supported by visible or spoken evidence to remove noise and duplicates.  Finally, it merges semantically related or consecutive actions into concise, instruction-oriented steps whose time ranges span from the earliest to the latest supporting subtitle to enforce temporal coherence and narrative conciseness.  This prompt-driven protocol yields a chronologically accurate and pedagogically clear sequence, continuously aligning linguistic cues with visual segments and stitching fragmented actions into unified, educator-friendly instructions.

% \textbf{Dataset Statistics.}

\section{Datasets \& Evaluation Metrics} \label{sec:Datasets Evaluation Metrics}

In this section, we detail each dataset and corresponding
evaluation metric.

\begin{figure*}[th]
    \centering
    \includegraphics[width=0.95\textwidth]{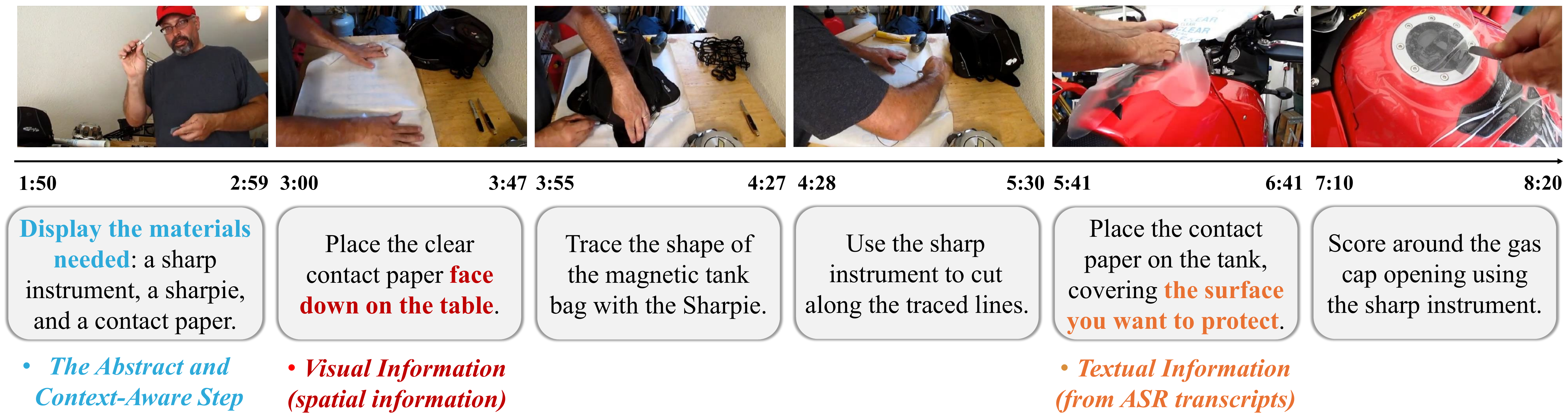}
    \caption{\textbf{Illustrative examples of our DenseCaption100 benchmark.} Annotated instructional steps show detailed instructions containing visual content and key ASR-derived textual information.}
    \label{fig:DenseCaption100}
\end{figure*}

\vspace{3pt}\noindent \textbf{DenseCaption100.} To facilitate a rigorous assessment of multimodal integration between ASR transcripts and visual cues, we curated the \textbf{DenseCaption100} benchmark. This dataset comprises manual annotations for 100 high-quality instructional videos sampled from the HowTo100M corpus, encompassing 12 diverse functional topics. Each video, with an average duration of 247.5 seconds, is densely annotated with an average of 11.0 detailed procedural steps per video. As illustrated in Figures~\ref{fig:DenseCaption100}, our annotation procedure captures both visual nuances and critical textual attributes derived from ASR transcripts—such as specific completion states or ingredient properties (e.g., ``whisk the batter until smooth''). To measure performance on this benchmark, we employ the R1-Score for dense video captioning to assess semantic alignment via DeepSeek-R1 matching, while for procedural step grounding, we report Intersection over Union (IoU) at multiple thresholds to evaluate temporal localization precision.

To ensure the reliability and consistency of the human-written annotations, we conducted an inter-annotator agreement (IAA) analysis on a randomly selected subset of 20 videos. Each video was independently annotated by two trained annotators following the same rigorous guidelines. We evaluated the agreement by treating one annotator's output as the reference and the other's as the prediction. After performing step-level matching based on temporal alignment, we measured the similarity of matched descriptions using SODA-c~\cite{fujita2020soda}, METEOR~\cite{banerjee2005meteor}, and our proposed R1-Score. 

As summarized in Table~\ref{tab:IAA}, the results indicate a high level of agreement between annotators. Specifically, the \textbf{R1-F1 score of 86.78\%} suggests that the vast majority of annotated steps are semantically aligned. Furthermore, the \textbf{mean IoU of 71.30\%} indicates substantial overlap in the temporal boundaries defined by different annotators, confirming the precision of our benchmark's temporal grounding. \textbf{All the data are released to the community.}

\begin{table}[ht]
\centering
\small
\caption{\textbf{Inter-Annotator Agreement (IAA) results on DenseCaption100.} We evaluate semantic and temporal consistency across 20 randomly selected videos.}
\label{tab:IAA}
\begin{tabular}{lcccccc}
\toprule
Metric & SODA-c $\uparrow$ & METEOR $\uparrow$ & P $\uparrow$ & R $\uparrow$ & F1 $\uparrow$ & mIoU $\uparrow$ \\
\midrule
Value & 23.1 & 35.6 & 86.63 & 86.93 & 86.78 & 71.30 \\
\bottomrule
\end{tabular}
\end{table}

\vspace{3pt}\noindent \textbf{YouCook2.} YouCook2 \cite{zhou2018towards} is a large-scale exocentric instructional dataset comprising 2,000 untrimmed videos across 89 cooking recipes. The dataset is characterized by narrations that describe fine-grained procedural actions. We leverage this benchmark to evaluate our model across two distinct dimensions:
\textbf{(i) Video-Text Retrieval}: To assess cross-modal alignment, we utilize the YouCook2 validation split consisting of 3,350 video clips with an average duration of 19.7 seconds.The model is required to retrieve the corresponding ground-truth caption for a given video clip from the entire pool of 3,350 candidates. Performance is quantified using Recall@1 (R@1) and Recall@5 (R@5).
\textbf{(ii) Dense Video Captioning}: To verify the generative quality of our automated pipeline, we evaluate its performance on the video-level validation set. This subset contains 436 long-form videos with an average duration of 212.4 seconds. We report SODA$_c$, CIDEr, and METEOR scores to measure the semantic and temporal accuracy of the generated instructions.

\vspace{3pt}\noindent \textbf{EgoMCQ.} To assess the model's capacity for fine-grained egocentric video-text alignment, we conduct experiments on the EgoMCQ benchmark~\cite{lin2022egocentric}. This task is formulated as a five-way multiple-choice problem: given a textual query, the model must identify the corresponding video clip from five potential candidates. Following the protocols in~\cite{lin2022egocentric}, we evaluate performance under two distinct settings: inter-video and intra-video. The latter is particularly rigorous, as all five candidates are sourced from the same original video, sharing nearly identical environments and actors, thereby demanding superior discriminative capabilities from the model. The evaluation utilizes the full EgoMCQ corpus derived from the Ego4d dataset~\cite{grauman2022ego4d}, comprising approximately 39k questions and 198k narrations. The dataset distribution consists of 61.5\% inter-video and 38.5\% intra-video samples. In line with established literature, we report Top-1 Accuracy as the primary performance metric to quantify the model's retrieval precision.

\vspace{3pt}\noindent \textbf{CharadesEgo.} To assess the model's capacity for cross perspective retrieval, we leverage the CharadesEgo dataset~\cite{sigurdsson2018charades}. While traditionally employed for action recognition across 157 daily activity categories, CharadesEgo is uniquely suited for perspective-invariant reasoning due to its paired egocentric and exocentric video structure. These pairs were captured through a successive recording protocol where actors first performed an action in a third-person view and subsequently replicated the same sequence while wearing a head-mounted camera. Following the methodology established by~\cite{xu2024retrieval}, we implemented a stringent filtering protocol to ensure high-fidelity evaluation and robust semantic alignment. Specifically, we only retained video-text pairs that achieved the maximum rating (Score = 7) for both visual quality and video-text relevance, as provided by the dataset metadata. This selection process resulted in a curated test set of 145 high-quality ego-exo video pairs, with an average duration of 22.7 seconds. Adhering to standard retrieval evaluation protocols, we report the Recall@1 to quantify the performance of cross-perspective matching.

\vspace{3pt}\noindent \textbf{MSRVTT.} MSRVTT~\cite{xu2016msr} is a cornerstone benchmark for open-domain video-text understanding, providing a diverse collection of 10,000 video segments paired with 200,000 descriptive captions. The dataset is characterized by its broad semantic coverage, encompassing a vast array of categories such as human daily activities, competitive sports, and intricate natural landscapes. For our retrieval experiments, we adhere to the standard "1K-A" split, which evaluates the model on a curated subset of 1,000 video-text pairs. Following established literature, we report the Recall@1 to quantify the model's zero-shot cross-modal matching capability across these heterogeneous visual scenes.

\vspace{3pt}\noindent \textbf{DiDeMo.} DiDeMo~\cite{anne2017localizing} is a benchmark designed for localized video-text grounding, consisting of 10,000 unedited, long-form videos sourced from Flickr. Each video is annotated with approximately four short sentences, which are meticulously localized and presented in their natural temporal order. Following established evaluation protocols, we focus on the paragraph-to-video retrieval task, where these localized sentences are concatenated into a single, comprehensive paragraph to represent the entire video's narrative. For our experiments, we adhere to the official test split, which comprises 1,004 videos and 1,004 concatenated captions (aggregated from 4,021 original short descriptions). We report the Recall@1 to quantify the model's capability in aligning multi-event textual descriptions with long-range visual sequences.

\vspace{3pt}\noindent \textbf{R1-Score Metric.} To specifically address the unique nuances of procedural instructions—where a single human-annotated action may correspond to multiple fragmented model outputs—we introduce the \textbf{R1-Score}. This metric leverages DeepSeek-R1~\cite{guo2025deepseek} to perform semantic matching between predicted steps and ground truth (GT) annotations. We define \textbf{R1-Precision} ($P$), \textbf{R1-Recall} ($R$), and the \textbf{R1-F1-score} ($F1$) as follows:
\begin{equation}
P = \frac{|\mathcal{M}_{\text{pred}}|}{|\mathcal{Y}_{\text{pred}}|}, \quad R = \frac{|\mathcal{M}_{\text{gt}}|}{|\mathcal{Y}_{\text{gt}}|}, \quad F1 = \frac{2 P R}{P + R}
\end{equation}
where $\mathcal{Y}_{\text{gt}}$ and $\mathcal{Y}_{\text{pred}}$ denote the sets of GT and predicted steps, respectively. $\mathcal{M}_{\text{gt}} = \{ y_i \in \mathcal{Y}_{\text{gt}} \mid \exists y_j \in \mathcal{Y}_{\text{pred}}, \phi_{\text{llm}}(y_i, y_j) = 1 \}$ represents the set of \textbf{matched} GT steps, and $\mathcal{M}_{\text{pred}}$ is defined analogously. The function $\phi_{\text{llm}}(\cdot) = 1$ denotes a semantic match verified by DeepSeek-R1. Temporal accuracy is further measured via \textbf{R1-mIoU}, defined as the mean Intersection-over-Union across all matched pairs.

\begin{figure*}[th]
    \centering
    \includegraphics[width=0.8\textwidth]{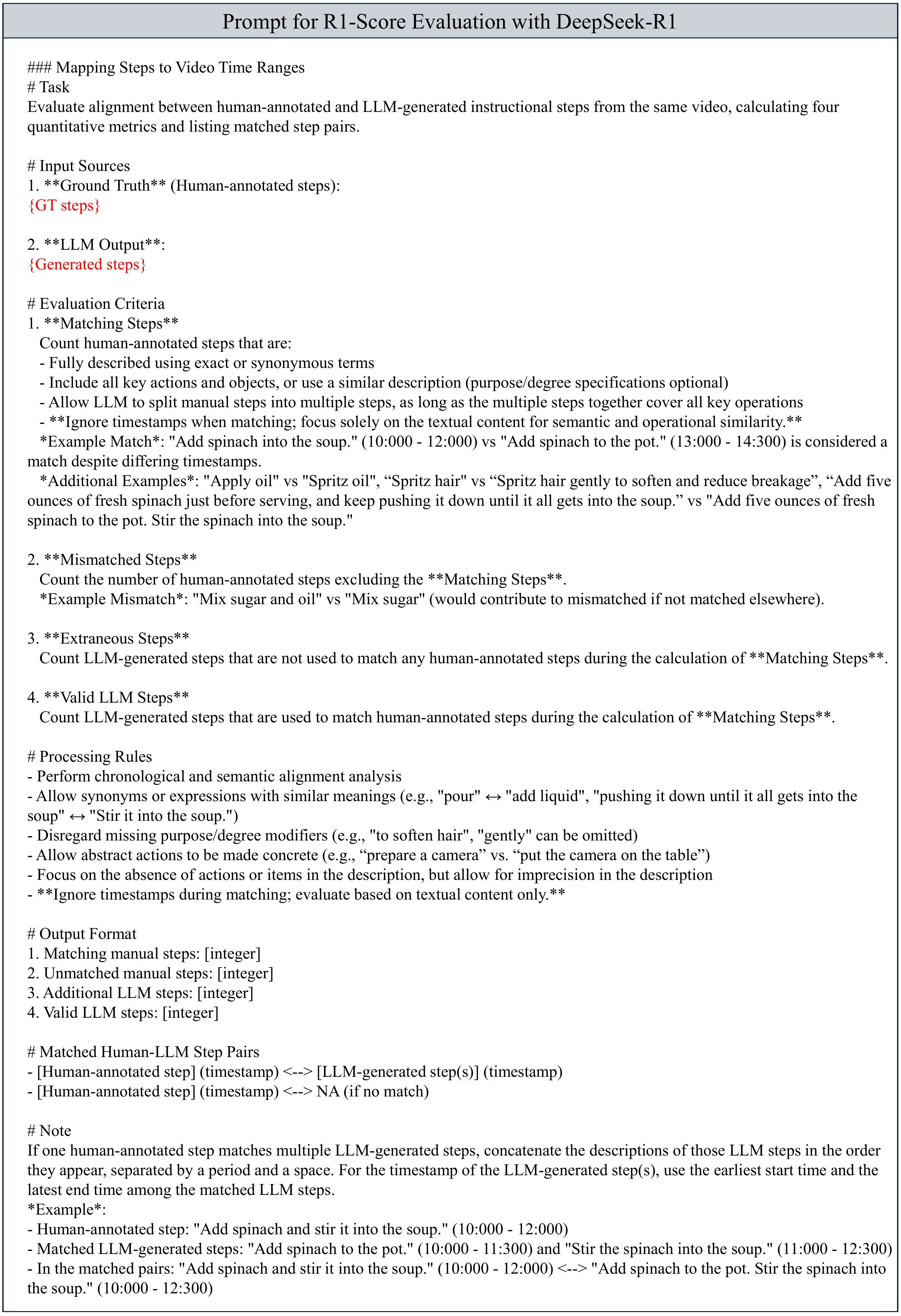}
    \caption{\textbf{Prompts for evaluating the R1-Score with DeepSeek-R1.} The full prompts can be found in our code.}
    \label{fig:Prompt_R1Score}
\end{figure*}

To compute these metrics, we construct a structured prompt for DeepSeek-R1~\cite{guo2025deepseek} (see Figure~\ref{fig:Prompt_R1Score}) that presents the GT and LLM-generated sequences side by side. Specifically, we instruct the model to:

\begin{itemize}[leftmargin=1em, itemindent=0pt]
    \item \textbf{Match manual and generated steps} by pairing each human-annotated action with one or more LLM steps whose combined descriptions \textbf{fully cover its key operations and objects}. We permit synonymous phrasing ({\em e.g.}, ``spritz oil'' $\leftrightarrow$ ``add oil''), abstract-to-concrete conversions ({\em e.g.}, ``prepare a camera'' $\leftrightarrow$ ``place the camera on the table''), and the omission of optional modifiers (e.g., adverbs of degree or purpose clauses).
    \item \textbf{Merge timestamps} for any GT step matched to multiple LLM steps, setting the start time to the earliest generated start and the end time to the latest generated end.
    \item \textbf{Compute R1 metrics}, including R1-Precision (the fraction of LLM steps that successfully match a human step), R1-Recall (the fraction of human steps that are matched by the LLM), R1-F1-score (the harmonic mean of R1-Precision and R1-Recall), and R1-mIoU (the mean Intersection-over-Union of timestamps across all matched pairs).
\end{itemize}

\begin{figure*}[th]
    \centering
    \includegraphics[width=0.8\textwidth]{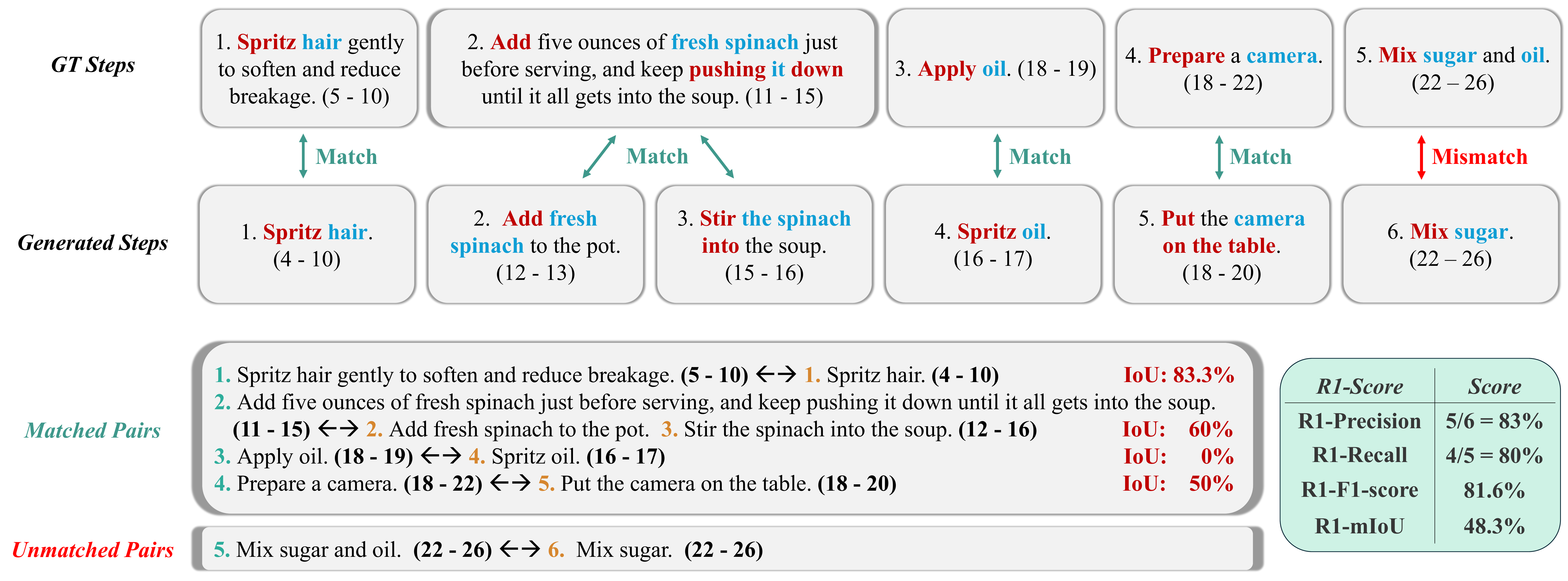}
    \caption{\textbf{Illustrative R1-Score computation for instructional step alignment.}  This example shows how we (1) \textbf{match each human-annotated action} with one or more LLM-generated steps whose combined descriptions fully cover its key operations and objects—allowing synonymous phrasing (e.g., “spritz oil” $ \leftrightarrow $ “add oil”), abstract-to-concrete conversions (e.g., “prepare a camera” $ \leftrightarrow $ “place the camera on the table”), step splitting or merging, and the omission of optional modifiers—and (2) \textbf{merge timestamps} for any Ground Truth step matched to multiple LLM steps by setting the start time to the earliest generated start and the end time to the latest generated end. From these aligned pairs, we then derive the R1 metrics (including Precision, Recall, F1-score and mIoU).}
    \label{fig:Examples_R1Score}
\end{figure*}

Unlike previous metrics such as SODA\_c~\cite{fujita2020soda}, the R1-Score allows a \textbf{single GT step} to match \textbf{multiple predicted steps}, prioritizing semantic coverage over one-to-one mapping. A set of predicted steps $P = \{p_1, \dots, p_n\}$ is considered a match for a GT step $g$ if: (i) $P$ encompasses all key operations/objects in $g$; and (ii) $P$ is a \textbf{minimal set}, such that removing any $p_i$ results in incomplete coverage. An illustrative computation of this alignment and the resulting metrics is shown in Figure~\ref{fig:Examples_R1Score}.

\section{Experiment Details} \label{sec:Implementation Details}

\begin{figure*}[th]
    \centering
    \includegraphics[width=0.8\textwidth]{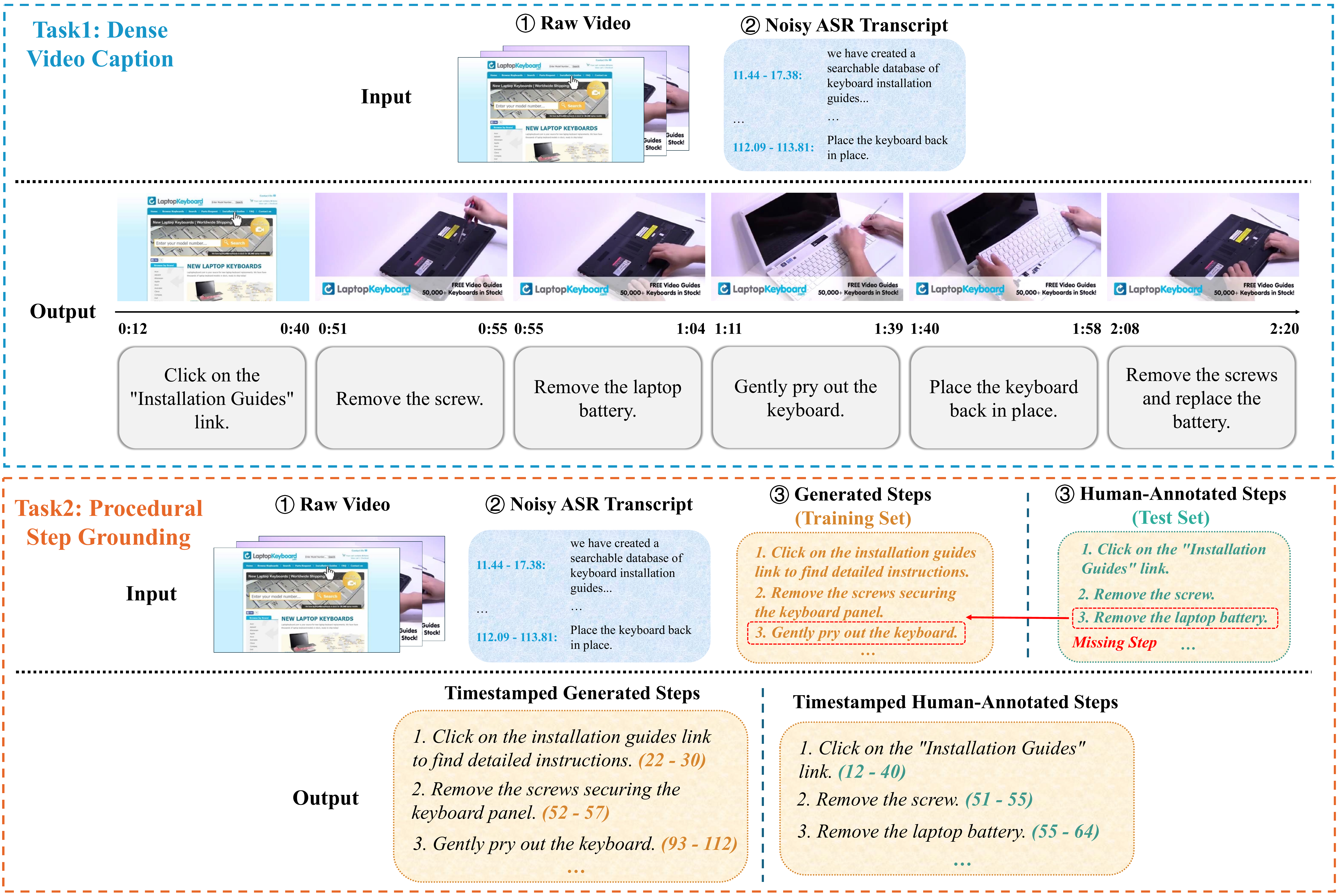}
    \caption{\textbf{Illustration of Downstream Tasks: Dense Video Captioning and Procedural Step Grounding.}
(a) \textbf{Dense Video Captioning}: the model generates timestamped, natural-language descriptions for every event in an instructional video.
(b) \textbf{Procedural Step Grounding}: the model localizes each procedural step within its corresponding video segment. When fine-tuned on the DenseStep2M training set, individual steps may be out of order, redundant, or missing; when evaluated on the DenseCaption100 test set, all steps are human-annotated and perfectly ordered.}
    \label{fig:Tasks_details}
\end{figure*}

\textbf{Tasks Details: Captioning and Grounding.} To evaluate the practical utility and transferability of DenseStep2M, we fine-tune \textbf{Qwen2.5-VL-3B}~\cite{bai2025qwen2} on two representative downstream tasks. First, in \textbf{Dense Video Captioning}, we train the model to produce timestamped, natural-language descriptions for every event in an instructional video, thereby generating a rich, sequential summary of its content (see ``Task1'' in Figure~\ref{fig:Tasks_details}). Second, in \textbf{Procedural Step Grounding}, we adapt the model to localize each procedural instruction within its corresponding video segment. Here, we fine-tune on our automatically curated DenseStep2M dataset—where occasional out-of-sequence, redundant, or missing steps—and evaluate on the human-annotated DenseCaption100 corpus, in which all steps appear in canonical order (see ``Task2'' in Figure~\ref{fig:Tasks_details}). This split allows us to measure both the model’s robustness to procedural irregularities and its ability to generalize to perfectly ordered instructions. 

\vspace{3pt}\noindent \textbf{Implementation Details: Captioning and Grounding.} We evaluated the effectiveness of our DenseStep2M dataset by fine-tuning the Qwen2.5-VL-3B model. During fine-tuning, we sampled up to 50 frames per video at 1 FPS, resizing them to 224×224 pixels. Both the visual encoder and the text decoder were jointly fine-tuned using LoRA adapters~\cite{hu2022lora}
on four Ascend 910B3 NPUs (256 GB total video memory, 64 GB per NPU). We used a batch size of 8 videos (with per\_device\_train\_batch\_size=1 and gradient\_accumulation\_steps=2), training for 1 epoch (approximately 6 hours). To optimize memory usage, we employed DeepSpeed's Zero Stage 3~\cite{rasley2020deepspeed}. The dataset was split into training and validation sets with a 99:1 ratio. Fine-tuning was conducted using the AdamW optimizer~\cite{loshchilov2017decoupled} with a learning rate of $1 \times 10^{-4}$. For inference, we extracted up to 100 frames per video (1 FPS, resolution 436×250) to capture longer temporal context.

\vspace{3pt}\noindent \textbf{Implementation Details: Cross-Modal Retrieval.} We evaluate two model variants for the cross-modal retrieval task: \textbf{Qwen3-VL-Embedding-8B} and \textbf{Qwen3-VL-8B-Instruct}. All retrieval experiments are conducted on a cluster of 8 $\times$ NVIDIA A100 (80GB) GPUs using DeepSpeed Stage 2 and BFloat16 mixed-precision training to optimize memory distribution.

For the \textbf{Embedding model}, we apply LoRA adapters ($r=32, \alpha=32$) to all linear layers within the vision tower and language model. Videos are sampled at a maximum of 8 frames ($256 \times 256$ pixels). To enhance discriminative power, we implement a two-stage pipeline featuring global hard negative mining on the DenseStep2M dataset. For each query, we retrieve the top-15 candidates and randomly sample 7 hard negatives whose similarity scores are at least 0.05 lower than the positive pair. The model is optimized for 1 epochs using a Symmetric Contrastive Loss ($\tau=0.05$) with the AdamW optimizer, a peak learning rate of $1 \times 10^{-6}$, and a 0.2-epoch linear warmup. To prevent out-of-memory errors, we employ gradient checkpointing and a chunked encoding strategy during the forward pass.

For the \textbf{Instruct model}, we employ a higher LoRA rank ($r=128, \alpha=256$) to fine-tune the projection and MLP layers. Each video is represented by 8 frames sampled at 8 FPS, with each frame processed into at most 64 patches (patch size 32) to maintain high throughput. We utilize a \textbf{Distributed Similarity Batch Sampler} to construct a competitive negative pool; 80\% of the samples in a batch are selected based on semantic similarity (similarity margin $>0.5$) using a pre-computed FAISS index~\cite{douze2024faiss}. The model is optimized via a Symmetric Masked InfoNCE loss ($\tau=0.03$), where an explicit masking mechanism excludes pairs with text similarity scores $\eta > 0.85$ to handle potential false negatives. Training is conducted for 1 epochs with a peak learning rate of $4.0 \times 10^{-5}$ and a global batch size of 768. We further enable Flash Attention 2~\cite{dao2023flashattention2} to maximize computational efficiency.

\vspace{3pt}\noindent \textbf{Implementation Details: Pipeline.} We employ a three‐stage pipeline to segment videos, generate instructional titles and steps, and refine the final output. First, we leverage Qwen2.5-72B~\cite{qwen2.5} to perform video segmentation and synthesize concise shot titles; this stage runs on four Ascend 910B3 NPUs (100‐way concurrency) with temperature = 0.7, top\_p = 0.8, top\_k = 20, repetition\_penalty = 1.05, max\_tokens = 1024, and a context window of 32,768 tokens. Next, Qwen2.5-VL-72B~\cite{bai2025qwen2} generates on-camera instructional steps by sampling frames at 1 FPS (up to 80 frames per shot) and distributing inference across two clusters of eight Ascend 910B3 NPUs each (10‐way concurrency), achieving a throughput of  6,000 shots/day (average shot length: 50 s) within a 32,768-token context. Finally, DeepSeek-R1-671B~\cite{guo2025deepseek} refines and integrates the steps using the same NPU configuration, with 16-way concurrency and a decoding speed of 275 tokens/s, processing 3,000 videos per day; we set max\_tokens = 16,384, temperature = 0.6, and retain the 32,768-token context window.

\begin{figure*}[th]
    \centering
    \includegraphics[width=0.85\textwidth]{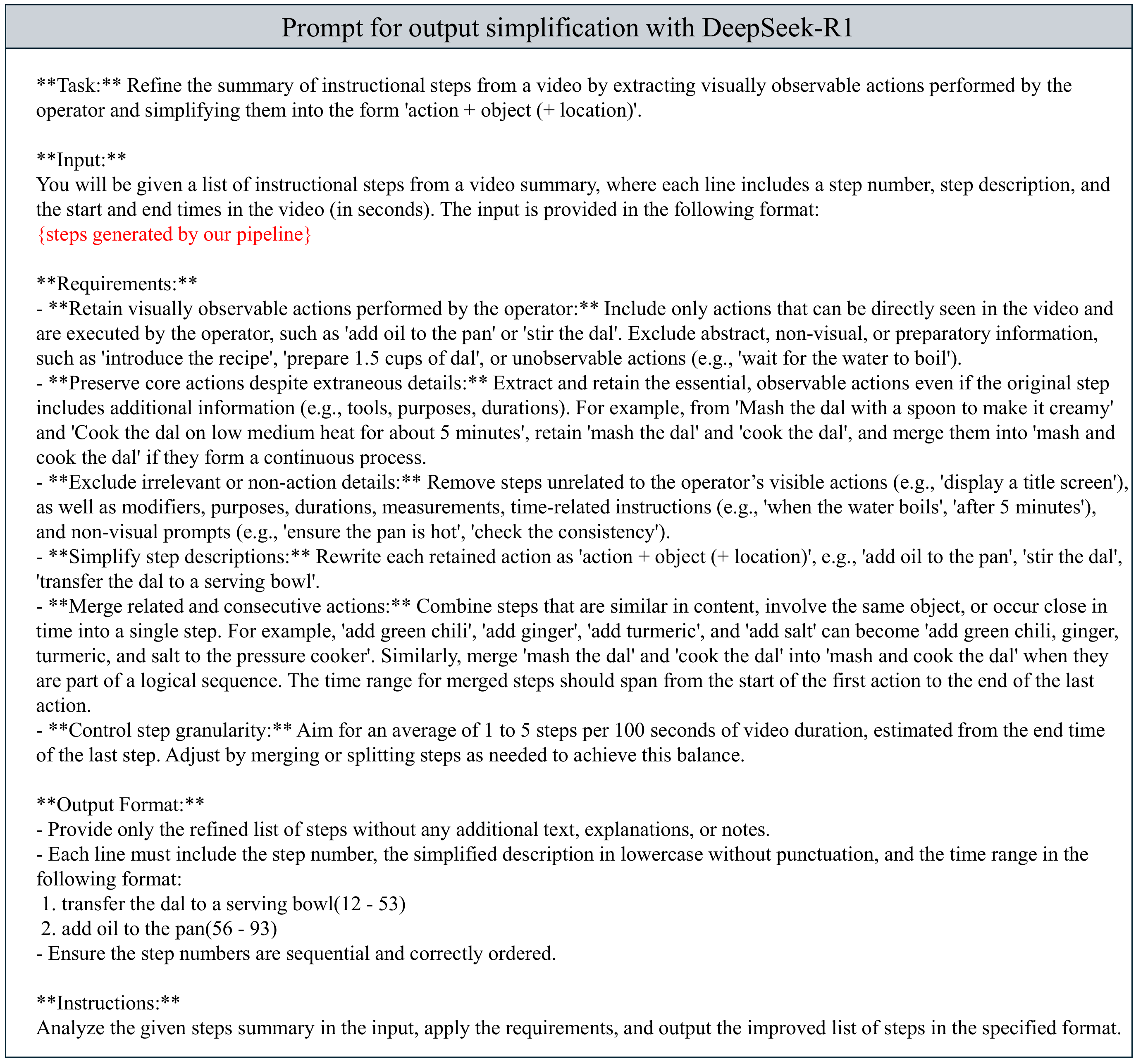}
    \caption{\textbf{Prompts for simplifying the steps generated by our pipeline to align with YouCook2 annotations~\cite{zhou2018towards} with DeepSeek-R1.} The full prompts can be found in our code.}
    \label{fig:Prompt_Youcook2}
\end{figure*}

\vspace{3pt}\noindent \textbf{Evaluation Details on YouCook2.} We observe that our pipeline produces substantially more detailed, fine‐grained steps than the original YouCook2 annotations~\cite{zhou2018towards} (see Sec.~\ref{subsec:Dataset Analysis}
in the main paper). To enable a fair comparison, we use DeepSeek-R1-671B~\cite{guo2025deepseek}—without any extra inputs (e.g., ASR transcripts)—to simplify our output via three operations: (i) remove modifiers, purposes, durations, measurements, and extraneous prompts, rewriting each step as a concise “action + object (+ location)” phrase; (ii) drop abstract or context‐dependent steps as described in Sec. ~\ref{subsec:Dataset Analysis}
; and (iii) merge semantically related steps to align with YouCook2’s annotation granularity. The full prompt guiding this process is shown in Figure~\ref{fig:Prompt_Youcook2}.

\section{Additional Quantitative Results} \label{sec:Additional Quantitative Results}

To assess the quality of the DenseStep2M dataset, we randomly sampled 10 videos from diverse categories, comprising 408 sentences in HowTo100M~\cite{miech2019howto100m} and 205 sentences in DenseStep2M.  As shown in Table \ref{tab:Manual Check}, DenseStep2M significantly improves both visual and temporal alignment compared to HowTo100M~\cite{miech2019howto100m}. 

\begin{table}[thbp]
\centering
\caption{\textbf{Manual check of dataset quality.} We manually evaluated two aspects: \textbf{(i)} the proportion of steps that are visually alignable with the video content, and \textbf{(ii)} the proportion of steps that are accurately aligned with their temporal boundaries.}
\label{tab:Manual Check}
\begin{tabular}{lcc}
\toprule
Dataset & Alignable \% $\uparrow$ & Well-aligned \% $\uparrow$ \\
\midrule
HowTo100M~\cite{miech2019howto100m} & 21.3 & 15.2 \\
DenseStep2M & \textbf{62.0} & \textbf{56.1} \\
\bottomrule
\end{tabular}
\end{table}

\section{Additional Qualitative Results} \label{sec:Additional Qualitative Results}

\begin{figure*}[th]
    \centering
    
    \begin{subfigure}[b]{0.8\textwidth}
        \centering
        \includegraphics[width=\textwidth]{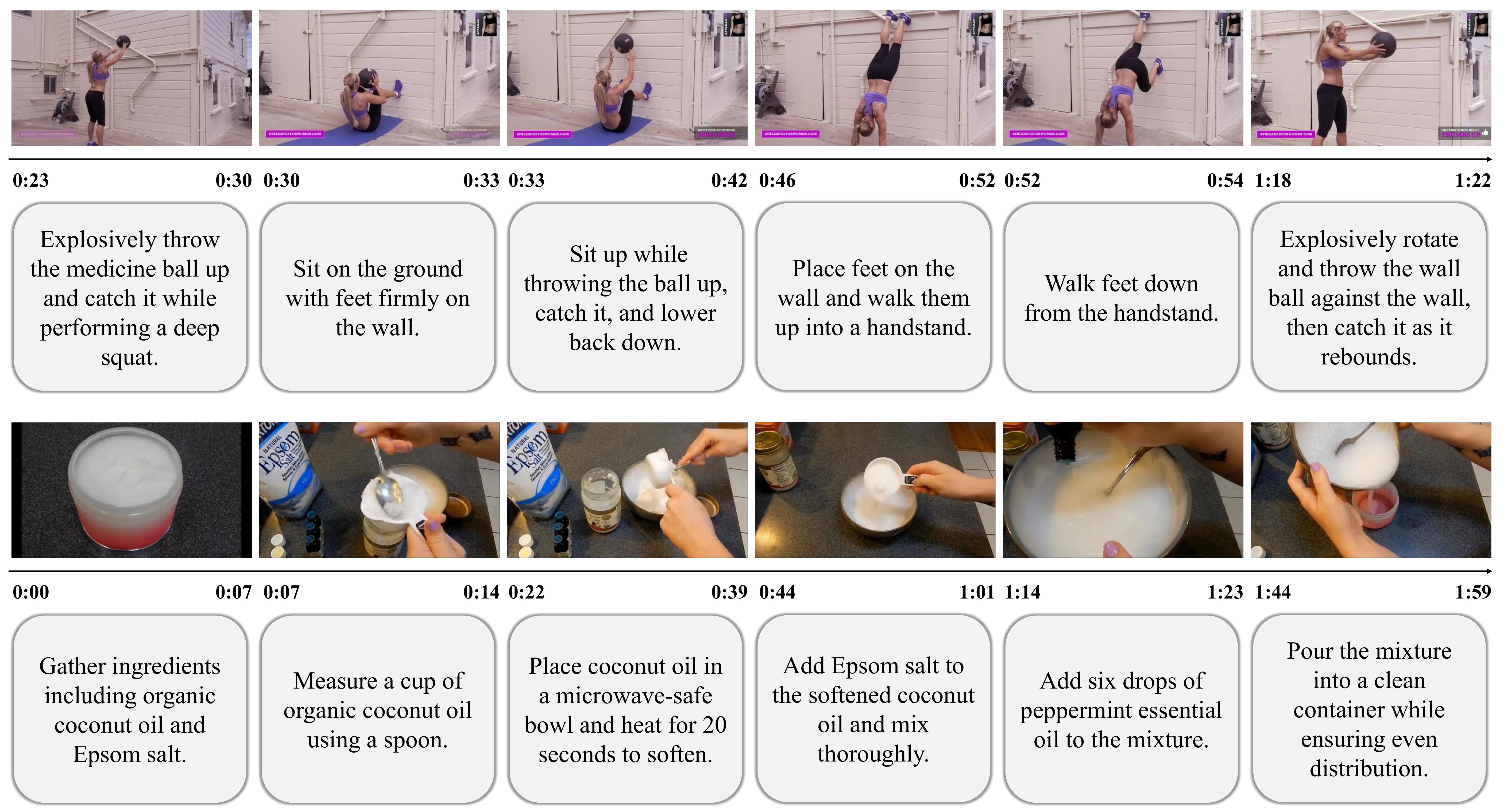}
        \caption{\textbf{Qualitative examples of our method on short videos.} Our pipeline successfully identifies and segments instructional steps with accurate timestamps.}
        \label{fig:QualitativeResults1}
    \end{subfigure}
    
    \vspace{0.3cm}
    
    \begin{subfigure}[b]{0.8\textwidth}
        \centering
        \includegraphics[width=\textwidth]{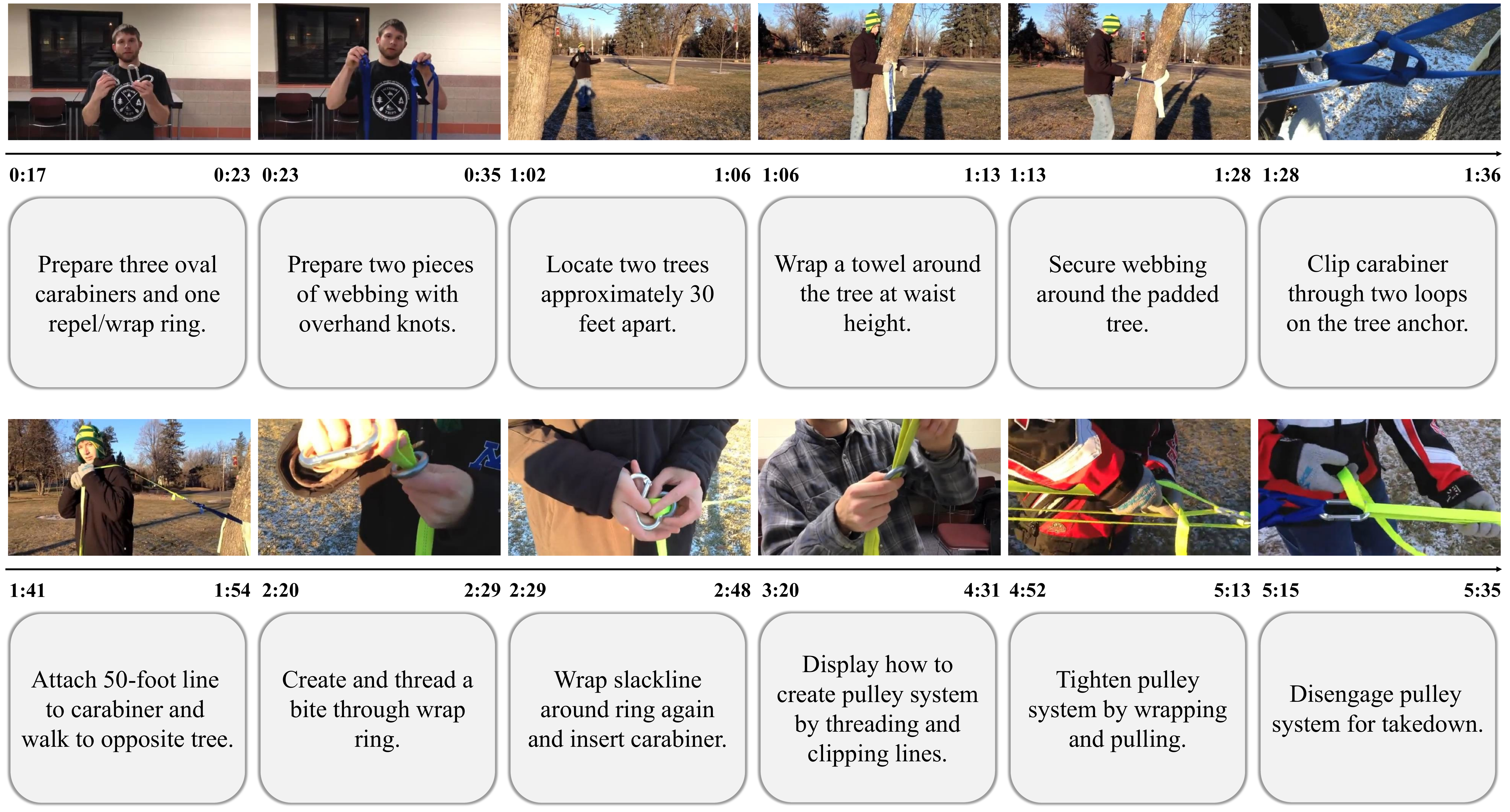}
        \caption{\textbf{Qualitative examples of our method on long videos.} The model maintains performance on longer temporal sequences, preserving step coherence and temporal alignment.}
        \label{fig:QualitativeResults2}
    \end{subfigure}
    
    \caption{\textbf{Additional qualitative results} demonstrating our model’s ability to produce temporally aligned instructional steps from diverse video inputs.}
    \label{fig:QualitativeResults}
\end{figure*}

We present additional qualitative results in Figure~\ref{fig:QualitativeResults}, illustrating the effectiveness of our pipeline in generating timestamped instructional sequences from input videos and textual prompts. These results demonstrate the robustness of our approach across both short and long video scenarios.

\section{Limitations and Ethical Concerns} \label{sec:Limitations and Ethical Concerns}
While DenseStep2M offers a scalable and high-quality approach to annotating instructional videos, it has several limitations. First, the pipeline relies heavily on the accuracy of ASR and multimodal LLMs, which can introduce errors or hallucinated steps, particularly in ambiguous or visually complex segments. Second, although visual-textual alignment is enforced, some misalignments may persist due to limitations in current model capabilities.  From an ethical standpoint, the dataset may inherit biases from the source videos in HowTo100M~\cite{miech2019howto100m}, including demographic, cultural, or content-related imbalances.  These factors should be carefully considered in future iterations of the dataset and its applications.

\section{Broader Impacts} \label{sec:Broader Impacts}
This work introduces an automated pipeline for annotating instructional videos, enabling scalable and high-quality procedural step extraction without manual effort. While primarily foundational, it could indirectly benefit applications such as education, assistive technologies, and robotics by enhancing long-form video understanding. However, like other generative systems, it may also carry risks such as misuse in surveillance or the generation of misleading procedural content. Mitigation strategies include dataset transparency, limiting deployment to non-sensitive domains, and encouraging responsible use by downstream practitioners.

\end{document}